# Image content dependent semi-fragile watermarking with localized tamper detection

**Samira Hosseini,[a] Mojtaba Mahdavi,[a,*]**
[a]University of Isfahan, Faculty of Computer Engineering, Department of IT Engineering, Hezar Jerib Ave., Isfahan, Iran, 81746-73441

**Abstract**. Content-independent watermarks and block-wise independency can be considered as vulnerabilities in semi-fragile watermarking methods. In this paper to achieve the objectives of semi-fragile watermarking techniques, a method is proposed to not have the mentioned shortcomings. In the proposed method, the watermark is generated by relying on image content and a key. Furthermore, the embedding scheme causes the watermarked blocks to become dependent on each other, using a key. In the embedding phase, the image is partitioned into non-overlapping blocks. In order to detect and separate the different types of attacks more precisely, the proposed method embeds three copies of each watermark bit into LWT coefficients of each 4×4 block. In the authentication phase, by voting between the extracted bits the error maps are created; these maps indicate image authenticity and reveal the modified regions. Also, in order to automate the authentication, the images are classified into four categories using seven features. Classification accuracy in the experiments is 97.97 percent. It is noted that our experiments demonstrate that the proposed method is robust against JPEG compression and is competitive with a state-of-the-art semi-fragile watermarking method, in terms of robustness and semi-fragility.

***Mojtaba Mahdavi**, E-mail: m.mahdavi@eng.ui.ac.ir

## 1 Introduction

Widespread and affordable access to the Internet, smartphones, and other digital devices allow individuals to acquire and share images with millions of people at the click of a button. People are now able to make unlimited copies of digital information [1]. Therefore, image copyright infringement is exceedingly likely. As a security mechanism, digital watermarking may be used to prevent this infringement [2, 3].

Digital watermarking refers to a process whereby data are embedded into a digital signal such as images, audio clips, or video sequences [2]. The covert marker, inserted in the signal is known as a digital watermark and usually contains signal-related information [4].

Based on the robustness, watermarking methods are divided into three categories: (1) robust, (2) fragile, and (3) semi-fragile [5]. In the first group, embedded watermark is robust against any distorting operation and remains extractable and usable. Contrarily, fragile methods are extremely sensitive to signal modification and are typically used to protect signal integrity [5]. Some fragile methods are capable of recovering the tampered regions [6, 7, 8]. Semi-fragile methods can withstand certain modifications [5]: they are robust against attacks which maintain signal content while being fragile against attacks that aim to modify the meaning of the signal. Semi-fragile methods can detect intentional attacks and identify or even reconstruct the tampered regions [9].

Watermarking methods have received considerable research attention; however, attacks on these methods remain to be investigated in greater depth [10]. In general, an attack is a process which interferes with watermark detection or the information communicated by the watermark in the carrier signal. Attacks are either intentional or unintentional [11].

As in cryptography, different levels of attacks can be carried out against watermarking methods, depending on the information available to the adversary [12].

### *1.1 Related works*

In the following, several watermarking methods with content-dependent watermarks are discussed. Rosales-Roldan et al. [13], developed a method with the localization and recovery capability of the tampered regions. In this algorithm, the watermark is in fact a scrambled halftone version of the original image which is inserted as an approximation of the image using the DWT or DCT coefficients. In the authentication stage, the halftone image is extracted and regenerated, and the two images are compared according to the Structural Similarity Index (SSIM) criterion to detect the tampered region.

A semi-fragile watermarking algorithm is presented in [9]. The algorithm consists of two parts: (1) a content-dependent singular-value-based sequence watermark and (2) a content-independent watermark generated using a secret key. The final watermark (secure watermark) is obtained by performing a logical operation on the two preceding watermarks. In the authentication stage, a three-level process verifies image authenticity by regenerating the secure watermark and computing five measures proposed by the authors.

The researchers proposed a two-stage detection method [14] against a number of attacks including the Collage attack. In this method, for each 8×8 block of pixels, six bits of watermark data are generated using significant DCT coefficients. For each block, the watermark bits are then embedded in the quantized DCT coefficients of the same block and other blocks. In the authentication stage, the first level of detection concerns detecting the general tampered regions using the GTW (General Tampering Watermark). Moreover, in order to identify the Collage attack, the GTW and CAW (Collage Attack Watermark) are used to define several parameters; if the parameter values exceed a determined threshold, the second level of detection will be executed.

Lee et al. [15] proposed a semi-fragile watermarking scheme wherein the image is splitted into 16×16 blocks of pixels; for each block, a watermark is generated using the first-order statistical moment of the block and inserted in the mid-frequency band of DWT coefficients, in another block. In the authentication stage, the blocks are mapped to each other using a random sequence. For instance, the watermark $w_i^*$ generated from block $x_i$ is embedded into block $x_{i+1}$, in which $i$ is a random sequence of positions generated using a secret key. To identify the tampered region, two comparisons are performed: (1) the watermark generated from $x_i$ with the watermark extracted from $x_{i+1}$ and (2) the watermark generated from $x_{i-1}$with the watermark extracted from $x_i$.

Authors in [16] presented a semi-fragile watermarking scheme for protecting biometric images, faces in particular. To achieve authentication, a watermark is generated from the Singular Value Decomposition (SVD) coefficients, obtained from each image block. A second watermark, namely the information watermark, is generated from the Principal Component Analysis (PCA) coefficients. Thereafter, both watermarks are inserted into wavelet medium-frequency coefficients using a proposed quantization method. Thus, by verifying the embedded watermark (used for authentication), the image is authenticated while tampered regions are detected and recovered using the information watermark.

A semi-fragile watermarking method is proposed in [17]. The method applies one level DWT on the image and a watermark is generated using LH,HL and HH coefficients. In the embedding procedure the image is partitioned into 2×2 blocks. Then one level DWT is applied on each block and LL coefficient is adjusted by quantizing it. Finally the inverse DWT is applied on each block and the watermarked image is obtained. By increasing the value of the quantization patamer, in the authentication phase, the method can detect the tampered region more precisely; but the PSNR value will have a significant decrease.

Chen et al. [18] proposed a semi-fragile method which is robust against JPEG compression. In this method the image is divided into non-overlapping 8×8 blocks and a 11-bit watermark is generated from each block; the mentioned watermark consists of 6-bit DC code and 5-bit AC code. In the embedding phase, by modifying 7 middle frequency DCT coefficients, the 11-bit watermark is embedded into each block. In the authentication phase, by comparing the extracted and generated watermark from the received image, tampered regions will be specified.

A content dependent semi-fragile watermarking method has been proposed in [19]. In order to be robust against unintentional attacks and be fragile against intentional ones, the method uses Zernik moments and Sobel edge map to generate a watermark. By exploiting the Zernik moments, the attack type (intentional or unintentional) will be specified and if the attack is intentional, Sobel edge map can specify the tampered regions.

The method proposed by Chetan et al. [20] uses Discrete Curvelet Transform coefficients as the watermark. In order to authenticate the received image, the extracted and the generated watermarks are partitioned into blocks of equal size. Each two blocks (an extracted watermark block and a generated watermark block) are compared with each other based on a similarity factor. If the similarity between two blocks exceeds a threshold, the block is a tampered one.

In the remainder of this paper, we set out to propose a semi-fragile watermarking method which is not vulnerable against the attacks that exploit the weaknesses like block-wise independency or a content independent watermark (has been used in the embedding process)

## 2 Proposed Semi-fragile Watermarking Method

The proposed scheme is mainly characterized by three features: (1) a two-part content-dependent watermark; (2) block-wise dependency such that the watermarked blocks are dependent on each other; and (3) a new authentication mechanism. It should be noted that the watermark bits are embedded into LWT[1] coefficients. In the following subsections, watermark generation approach, embedding, extraction and authentication procedures are detailed.

[1] Lifting Wavelet Transform

*2.1 Content-dependent Watermark Generation and Embedding*

In order to address the shortcomings in block-wise independent schemes, a two-part content-dependent watermark is generated. The generation and embedding procedure takes advantage of a secret key $K = k_1|k_2|k_3$; each part of the key is used to generate pseudo-random numbers in different stages. Fig. 1 demonstrates the general process of watermark generation and embedding.

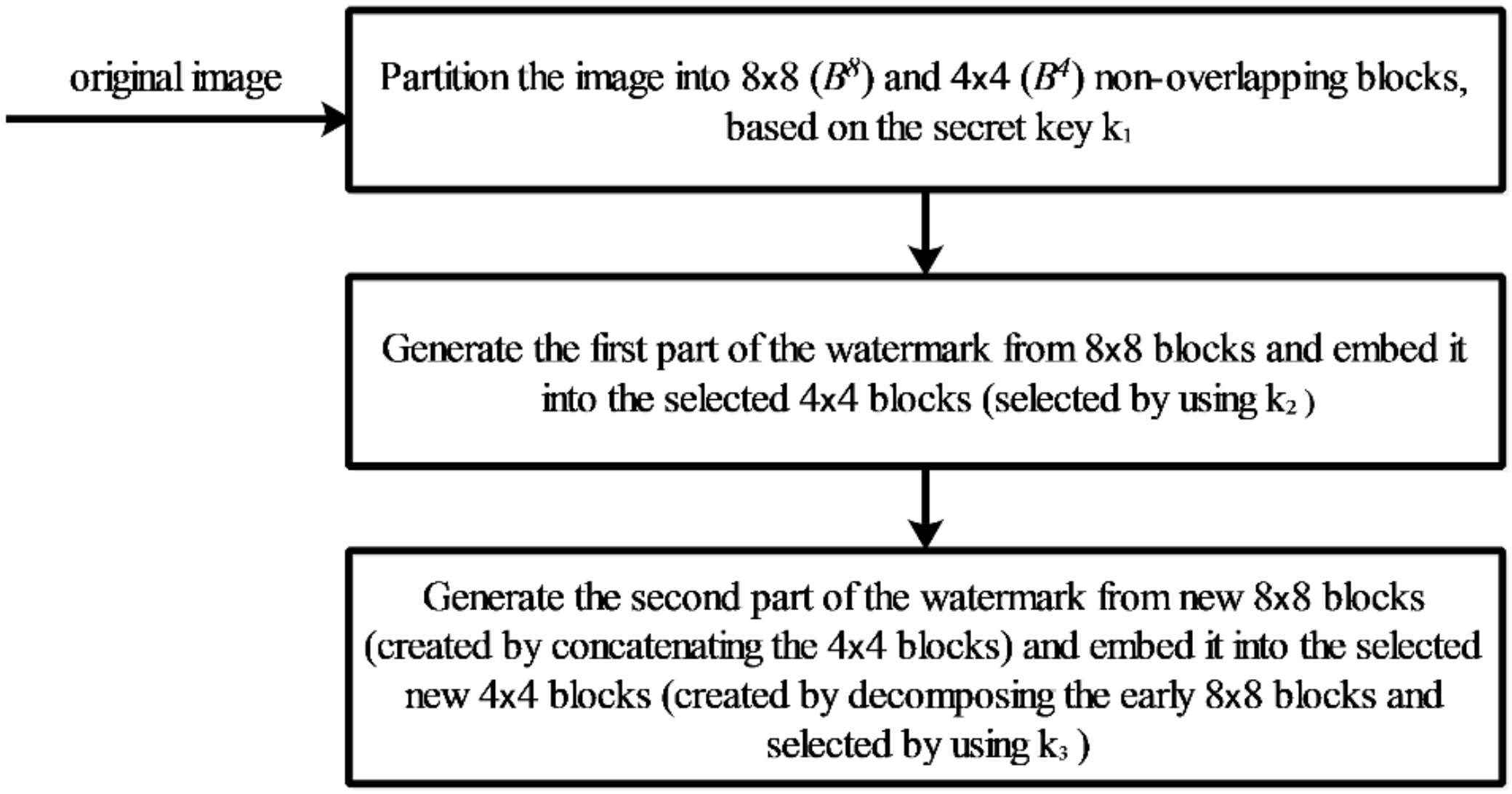

**Fig. 1** The process of watermark generation and embedding

Firstly, the image is split into multiple blocks. In doing so, uniformly distributed (pseudo)random non-overlapping $4 \times 4$ and $8 \times 8$ blocks are generated using the secret key $k_1$. The partitions are made such that a $4 \times 4$ block is four times as likely to be generated as an $8 \times 8$ block. Thus, assuming that the image is made up of $n$ blocks of $4 \times 4$, the required number of $4 \times 4$ blocks equals $n/2$ while $n/8$ blocks of $8 \times 8$ are required. To obtain these (pseudo)random blocks using the intended secret key, a sequence of (pseudo)random numbers is generated and (based on these (pseudo)random numbers) the partitions are created with minor changes of probabilities. As a result, the number of $4 \times 4$ blocks roughly equals four times the number of $8 \times 8$ blocks. Having $k_1$, the receiver is able to replicate the exact partitioning of the image.

Note that once each section of the image is partitioned, new blocks are selected and generated at the first possible location where none of the pixels are selected.

Fig. 2 illustrates an example of the proposed (pseudo)random partitioning.

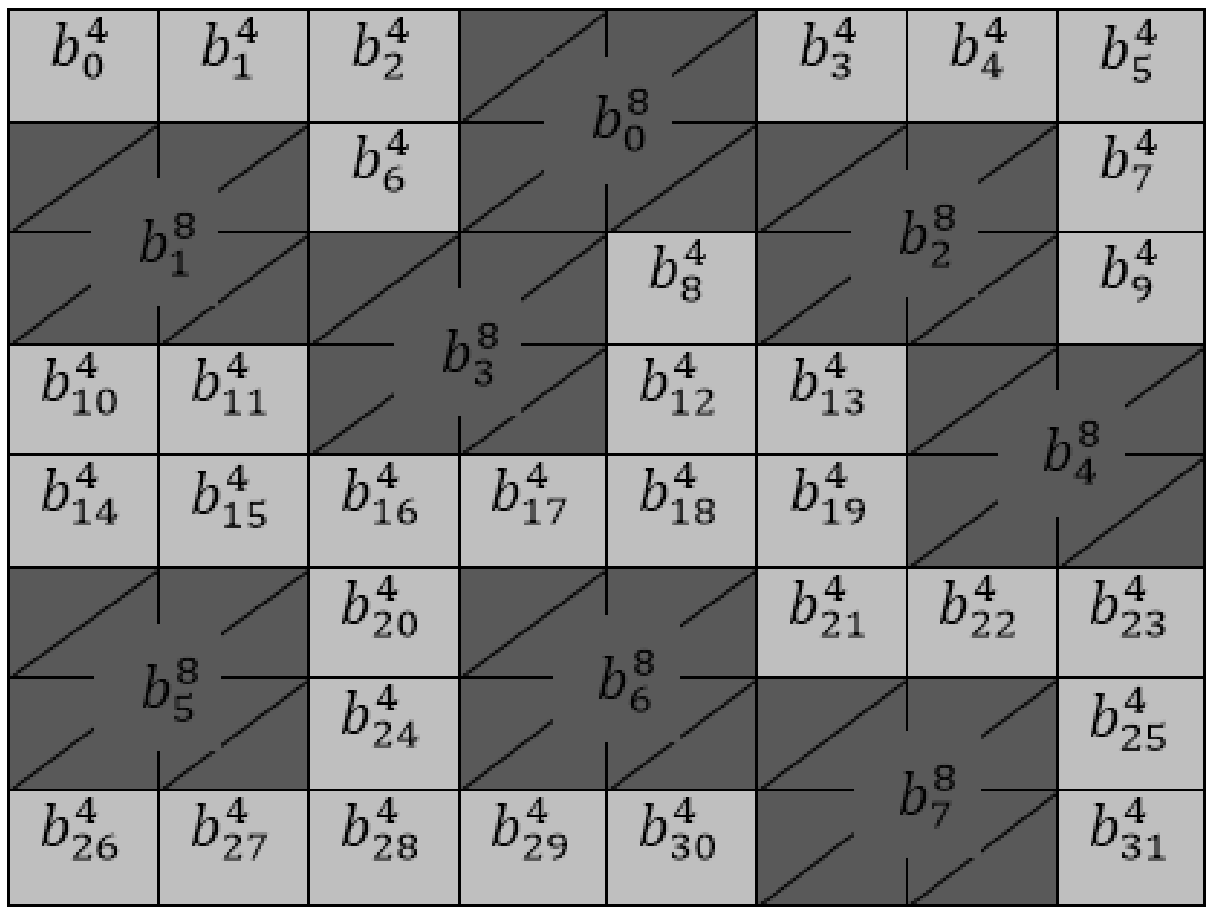

**Fig. 2** Random partitioning into $4 \times 4$ and $8 \times 8$ blocks

As shown, each cell represents a $4 \times 4$ block; memberships of the $B^8$ and $B^4$ sets are indicated for each block. In Eq. 1, the $8 \times 8$ and $4 \times 4$ blocks are denoted by $B^8$ and $B^4$, respectively.

$$B^8 = \{b_i^8 \mid 0 \le i < n/8\}$$
$$B^4 = \{b_j^4 \mid 0 \le j < n/2\} \quad (1)$$

As a result of this partitioning, the number of $4 \times 4$ blocks may be slightly (smaller than one percent) larger than the four times the number of $8 \times 8$ blocks. However, as demonstrated in our experiments, the difference is negligible.

Subsequent to partitioning, the $B^8$ blocks are used to generate the first part of the watermark. The following steps are taken to generate the watermark from each block:

1- Average pixel values for block $b_i^8$, denoted by $Avg_{b_i^8}$, is computed and the interval in which the value falls is identified according to Table 1.

2- Each interval from Table 1 corresponds to a four-bit gray code. Depending on the interval as well as the corresponding gray code, four bits are assigned to $b_i^8$. Indeed, four bits are generated from each $b_i^8$, using Eq. 2.

$$fourBits_{b_i^8} = grayCodeOf(\lfloor Avg_{b_i^8}/16 \rfloor) \tag{2}$$

It should be noted that two consecutive gray codes have only one different bit [21]. Thus, by using the gray code to generate four bits per block, the watermark becomes more robust against unintentional changes. Consider a scenario in which, as a result of unintended changes, the average value calculated by the receiver falls in the preceding or succeeding interval. The error involves only one bit while the remaining three bits are correct.

Thus, the first part of the watermark is created.

Now, each four bits of watermark generated from an $8 \times 8$ block, are embedded in four $4 \times 4$ blocks which are randomly selected using the secret key $k_2$. Considering $B'^4$ (in Eq. 3), four of the $4 \times 4$ blocks belonging to this set (e.g. $\{b_1'^4, \dots, b_4'^4\}$) are used to embed each four bits of the watermark.

$$B'^4 = permute_{k_2}(B^4) = \{b_j'^4 \mid 0 \le j < n/2\} \tag{3}$$

In Eq. 3, $permute_{k_2}(B^4)$, represents a (pseudo)random permutation of $B^4$ members based on $k_2$. It is noted that the embedding procedure (for a block) will be described in the next section.

**Table 1** Numeric intervals and the corresponding gray codes

| Interval | Corresponding Gray Code |
|---|---|
| [0,16) | 0000 |
| [16,32) | 0001 |
| [32,48) | 0011 |
| [48,64) | 0010 |
| [64,80) | 0110 |
| [80,96) | 0111 |
| [96,112) | 0101 |
| [112,128) | 0100 |
| [128,144) | 1100 |
| [144,160) | 1101 |
| [160,176) | 1111 |
| [176,192) | 1110 |
| [192,208) | 1010 |
| [208,224) | 1011 |
| [224,240) | 1001 |
| [240,255] | 1000 |

Next, $4 \times 4$ blocks of the image (members of $B^4$), which now contain the bits of the first part of the watermark, are used to generate the second part. By considering each $8 \times 8$ block as composed of four $4 \times 4$ blocks, one bit may be embedded in each $4 \times 4$ block. To create the second part of the watermark, $4 \times 4$ blocks in $B^4$ are sequentially converted into $8 \times 8$ blocks ($B'^8$). Thus, using the aforementioned steps, four watermark bits are generated using each $8 \times 8$ block (members of $B'^8$) and embedded in $4 \times 4$ blocks obtained from breaking down $8 \times 8$ blocks (members of $B^8$). Given the fact that close $4 \times 4$ blocks (blocks in $B^4$) are similar in color, they are used to create the second part of the watermark. Although the disparity in the colors may be large in some cases, the issue is not significant as confirmed by the experiments. In Eq. 4, by joining four consecutive members of $B^4$, members of $B'^8$ are generated.

$$b_i'^8 = \{b_j^4 \mid j \in \{i \times 4, \dots, (i \times 4) + 3\}, 0 \le j < n/2\}$$
$$B'^8 = \{b_i'^8 \mid 0 \le i < n/8\} \quad (4)$$

Each $8 \times 8$ block of the image contains four $4 \times 4$ blocks. The second part of the watermark is embedded in $4 \times 4$ blocks selected from $8 \times 8$ blocks (members of $B^8$). In fact, for the second part of the watermark, the four bits generated based on $8 \times 8$ blocks (members of $B'^8$) are embedded in four random blocks selected from $8 \times 8$ blocks of the original image (members of $B^8$) using $k_3$. In other words, using Eq. 5, the $4 \times 4$ blocks obtained from dividing $8 \times 8$ blocks are reordered using $k_3$ to create $B''^4$. Ultimately, members of $B''^4$ are used to embed the second part of the watermark according to the procedure will be explained in the next section.

$$b_i^8 = \{bl_{x,y}^4 | \; x \in \{0,1\}, y \in \{i \times 2, (i \times 2) + 1\}, 0 \leq i < n/8\} \tag{5}$$
$$B''^4 = permute_{k_3}\{bl_{x,y}^4 \big| \; x \in \{0,1\}, y \in \{i \times 2, (i \times 2) + 1\}, 0 \leq i < n/8\}$$

In summary, the watermark consists of two parts and involves two stages. Initially, the first part of the watermark is generated from $8 \times 8$ blocks and embedded in randomly selected $4 \times 4$ blocks. Then, the second part is created using close $4 \times 4$ blocks of the image (which are joined to create $8 \times 8$ blocks) and embedded in (randomly selected) $4 \times 4$ blocks contained in larger blocks.

By reaching this point, the watermark will depend on the content of the image. Furthermore, the watermark is embedded in a block-wise dependent fashion. As a result, a number of attacks including the attack in [22], Collage, and Vector quantization are rendered ineffective thus precluding damage to image authenticity. The next section explains the embedding procedure.

### *2.2 Watermark Embedding and Extraction*

Firstly, the watermark to be inserted into the image is generated based on the method, proposed in the previous section. Prior to embedding the watermark, the image is prepared in two steps: (1) to get the four subbands LL, HL, LH and HH, LWT is performed on each $4 \times 4$ block. The non-overlapping $8 \times 8$ and $4 \times 4$ blocks are obtained based on the procedure mentioned in the previous section; and (2) LWT is again performed on the LL, HL, and LH subbands and $LL_{LL}$, $LL_{HL}$, and $LL_{LH}$ coefficients are selected for embedding the watermark. It is noted that lifting wavelet transform is the second generation wavelet transform which consists of three steps: splitting , predicting and updating [23]. In comparison to DWT, LWT consumes less memory and reduces information loss [24].

The next step involves inserting each watermark bit in each block; three identical bits are embedded into $LL_{LL}$, $LL_{HL}$, and $LL_{LH}$ coefficients. In fact, for better separation of different types of attacks, the proposed method embeds three copies of each watermark bit into LWT coefficients of each 4×4 block. Eq. 6 shows how the watermark bits are embedded in the mentioned coefficients of a block.

$$y = \begin{cases} lsb(\left\lfloor \frac{coef}{q} \right\rfloor) \times q & lsb\left(\left\lfloor \frac{coef}{q} \right\rfloor\right) = w \\ lsb(\left\lfloor \frac{coef}{q} \right\rfloor + 1) \times q & otherwise \end{cases} \tag{6}$$

It is noted that *q* is the quantization step and *w* represents the watermark bit.

In order to extract the watermark bits, the proposed partitioning is performed, then LWT is applied on each $4 \times 4$ block and again it is applied on LL, HL, and LH subbands. Finally the bits are extracted from $LL_{LL}$, $LL_{HL}$, and $LL_{LH}$ coefficients using Eq. 7. in the following equation $\widetilde{w}$ represents the extracted bit from the corresponding coefficient (*coef*).

$$\widetilde{w} = lsb(round(\frac{coef}{q})) \tag{7}$$

*2.3 Watermark Authentication*

As the watermark is generated and embedded in two stages, watermark extraction and image authentication involve two steps. At first, the first part of the watermark is extracted and reconstructed. Given the differences between the reconstructed and extracted bits, four matrices namely $X_{w11}$, $X_{w12}$, $VMap_{11}$, and $VMap_{12}$ are created. Next, the second part of the watermark is extracted and reconstructed. As before, by comparing the extracted and reconstructed bits, the $X_{w21}$, $X_{w22}$, $VMap_{21}$, and $VMap_{22}$ matrices are created. To authenticate the image, a number of features are computed using the constructed matrices. These features allow us to classify images into one of four categories: clean watermarked images, unintentionally modified, intentionally tampered and intentionally and unintentionally modified images. Moreover, an error map is created using $X_{w11}$, $X_{w12}$, $X_{w21}$, and $X_{w22}$ to visually represent image authenticity. It also displays potential tampered regions. Fig. 3 shows the general process of watermark extraction and authentication.

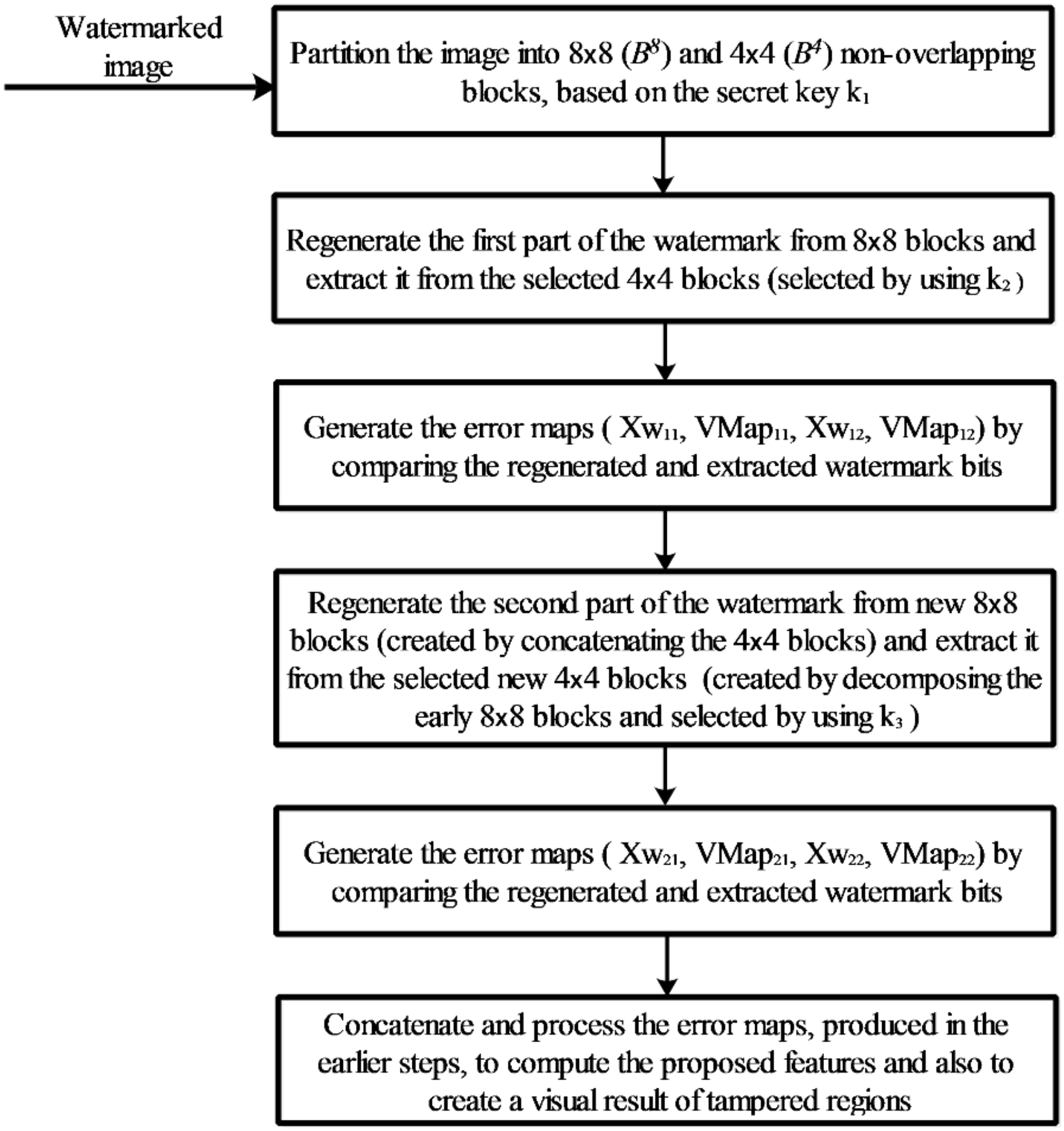

**Fig. 3** general process of watermark extraction and authentication

In the following, the steps to construct the aforementioned eight matrices, calculating the proposed features, and generating the error map (to identify tampered regions) are explained in full detail. The steps of watermark extraction and image authentication are as follows:

1- As before, the received image is split into non-overlapping $8 \times 8$ and $4 \times 4$ blocks ($B^8$ and $B^4$), using the secret key $k_1$.

2- The first part of the watermark is reconstructed from $8 \times 8$ blocks using the procedure in section 2.1.

3- Using the secret key $k_2$, $4 \times 4$ blocks (members of $B'^4$) containing the first part of the watermark are selected.

4- As discussed, the four bits generated from each $8 \times 8$ block are embedded into four randomly selected $4 \times 4$ blocks. Therefore, following the extraction of three bits from each $4 \times 4$ block using Eq. 7, the differences between a reconstructed watermark bit and the respective extracted bits are calculated, using Eq. 8 [25].

$$\begin{aligned} Ew(1) &= |w - \widetilde{w}(1)|, \\ Ew(2) &= |w - \widetilde{w}(2)|, \\ Ew(3) &= |w - \widetilde{w}(3)|. \end{aligned} \tag{8}$$

where $w$ denotes a reconstructed watermark bit (from an $8 \times 8$ block) and $\widetilde{w}(1)$, $\widetilde{w}(2)$ and $\widetilde{w}(3)$ represent the three extracted bits (from a $4 \times 4$ block).

5- Now, using $Ew(1)$, $Ew(2)$, and $Ew(3)$ [25] four matrices are computed: $X_{w11}$, $X_{w12}$, $VMap_{11}$, and $VMap_{12}$. In the proposed method, the watermark is dependent on image content; thus, $X_{w11}$ and $X_{w12}$ must be computed to represent the bits reconstructed from $8 \times 8$ blocks (the first part of the watermark) and the bits extracted (using Eq. 7 for the first part of the watermark) from $4 \times 4$ blocks of the watermarked image, respectively. Furthermore, $VMap_{11}$ represents the bits reconstructed from $8 \times 8$ blocks of the image (for the first part of the watermark) while $VMap_{12}$ represents the bits extracted from $4 \times 4$ blocks of the watermarked image (using Eq. 7 for the first part of the watermark).

6- $Ew(1)$ values for each block define $X_w$ elements. For each $4 \times 4$ block, $Ew(1)$ contains either 1 or 0. If $Ew(1) = 1$, then $X_w(m,n) = 1$ (or $X_w(m,n) = 255$) indicating that the extracted bit has been intentionally tampered; in contrast, if $Ew(1) = 0$ then $X_w(m,n) = 0$. The proposed method leverages the same procedure for $4 \times 4$ blocks from which the watermark bits are extracted to construct $X_{w12}$. With respect to $8 \times 8$ blocks, four values for $Ew(1)$ exist since four watermark bits are obtained from each block. The extent of tampering for each $8 \times 8$ block can be determined using the four obtained values (four values for $Ew(1)$). Let $Ew(1)_{b1}$, $Ew(1)_{b2}$, $Ew(1)_{b3}$, and $Ew(1)_{b4}$ represent the values obtained for the reconstructed bits from an $8 \times 8$ block, then the respective value in $X_{w11}$ is obtained as shown in Table 2.

**Table 2** $X_{w11}$ calculation procedure for each $8 \times 8$ block

| Sum ($Ew(1)_{b1}$, $Ew(1)_{b2}$, $Ew(1)_{b3}$, $Ew(1)_{b4}$) | The element representing an 8 ×8 block in $X_{w11}$ |
|---|---|
| 0 | 0 |
| 1 | 63 |
| 2 | 127 |
| 3 | 191 |
| 4 | 255 |

So far, $X_{w11}$ and $X_{w12}$ are obtained for the first part of the watermark (reconstructed bits and extracted bits).

7- Henceforth, $VMap_{11}$, and $VMap_{12}$ are calculated. Initially, $VMap$ is calculated using $Ew(.)$ which equals 0, 1, 2, or 3 for each $4 \times 4$ block. Table 3 presents possible $VMap$ values, computed based on $Ew(.)$.

**Table 3** $VMap$ calculation procedure for each $4 \times 4$ block [25]

| $Ew(.)$ | $VMap(m,n)$ |
|---|---|
| 0 0 0 | 0 |
| 0 0 1 | 1 |
| 0 1 0 | 1 |

| | | | |
|---|---|---|---|
| 0 | 1 | 1 | 1 |
| 1 | 0 | 0 | 2 |
| 1 | 1 | 0 | 2 |
| 1 | 0 | 1 | 2 |
| 1 | 1 | 1 | 3 |

The proposed method uses the same approach for $4 \times 4$ blocks from which the watermark bits are extracted to create $VMap_{12}$. Possible values for the elements are as follows: 0, 1 (or 85), 2 (or 170), and 3 (or 255). However, with respect to $8 \times 8$ blocks, for each reconstructed bit, three values are obtained i.e. $Ew(1)$, $Ew(2)$, and $Ew\ (3)$. Thus, for each bit, an integer value ranging from 0 to 3 is calculated using the three $Ew(.)$ values and Table 3. In other words, four values are obtained for the four bits reconstructed from an $8 \times 8$ block. To obtain a single value for each block, a voting process is required. The single value represents the element corresponding to a block in $VMap_{11}$. The voting algorithm is as follows:

- The number of times each of the values 0, 1, 2, or 3 occurs is determined. The counts are denoted by $c_0$ to $c_3$, respectively. The corresponding element in $VMap_{11}$ is calculated using the algorithm shown in Fig. 4.

```
If ((c3 + c2) ≥ (c1 + c0))
    If (c3 ≥ c2)
        VMap11 (block) = 255;
    Else
        VMap11 (block) = 170;
    End
Else
    If (c1 ≥ c0)
        VMap11 (block) = 85;
    Else
        VMap11 (block) = 0;
    End
end
```

**Fig. 4** Voting algorithm to construct $VMap_{11}$and obtain single values for each $8 \times 8$ block

As a result, the extent of tampering for an $8 \times 8$ block is determined using $VMap_{11}$. To obtain visible images, the elements of the matrices ($VMap_{12}$ and $VMap_{11}$) are multiplied by 85 (Fig. 2).

Now, the second part of the watermark is extracted and authenticated:

8- The $4 \times 4$ blocks in $B^4$ are sequentially selected and converted into new $8 \times 8$ blocks ($B'^8$). Then the second part of the watermark is reconstructed from these blocks.

9- Using $k_3$, the intended $4 \times 4$ blocks (members of $B''^4$) obtained by breaking down $8 \times 8$ blocks (generated in step 1) are selected. After extracting three bits from each $4 \times 4$ block (using Eq. 7), the differences between the reconstructed and extracted bits are calculated using Eq. 8.

10- Using $Ew(1)$, $Ew(2)$, and $Ew$ (3), the $X_{w21}$, $X_{w22}$, $VMap_{21}$, and $VMap_{22}$ matrices are computed. As mentioned in Step 5, in the proposed method, the watermark is dependent on the content of the image. Thus, it is necessary to construct four matrices: (1) $X_{w21}$ representing the reconstructed bits from $8 \times 8$ blocks ($B'^8$) (2) $X_{w22}$ showing the bits extracted from $4 \times 4$ blocks (members of $B''^4$) using Eq. 7 for the second part of the watermark; also (3) the $VMap_{21}$ and (4) $VMap_{22}$ matrices are calculated with the same purpose as $X_{w21}$ and $X_{w22}$.

11- To construct $X_{w22}$ (for the $4 \times 4$ blocks from which the watermark bits are extracted), the procedure in Step 6 is performed. Also, the relevant procedure from Step 6 is applied to $8 \times 8$ blocks created by joining $4 \times 4$ blocks from $B^4$. Thus, the operation yields $X_{w21}$.

12- $VMap_{22}$ is created using the procedure in Step 7 for the $4 \times 4$ blocks from which the watermark bits are extracted (members of $B''^4$). Furthermore, the procedure pertaining to $8 \times 8$ blocks in Step 7 is applied to $8 \times 8$ blocks obtained via joining $4 \times 4$ blocks from $B^4$. Following the voting operation, $VMap_{21}$ is obtained.

By following the preceding steps, a total of eight matrices are constructed: $X_{w11}$, $X_{w12}$, $VMap_{11}$, $VMap_{12}$ , $X_{\mathrm{w21}}$, $\mathrm{X}_{\mathrm{w22}}$, $VMap_{21}$, and $VMap_{22}$. Next, these matrices are processed to yield $X_{w1}$, $X_{w2}$, $VMap_1$, and $VMap_2$.

13- Since each element in $X_{w11}$ and $VMap_{11}$ represents an $8 \times 8$ block, each element is expanded to $8 \times 8$ blocks. In a similar manner, since each element of $X_{w12}$ and $VMap_{12}$ represent a $4 \times 4$ block, each element is expanded to $4 \times 4$ blocks. Now, the $X_{w1}$ and $VMap_1$ matrices with the same dimensions as the original image (i.e. $512 \times 512$) are defined and the elements are initialized to zero. In $X_{w1}$, expanded elements from $X_{w11}$ and $X_{w12}$ are assigned to the locations of their respective representative blocks in the original image. Thus, the $X_{w1}$ error map is formed. Using a similar approach, $VMap_1$ is constructed using the two expanded matrices $VMap_{11}$ and $VMap_{12}$. The function of these matrices is described in a later section. Note that the original location of each $4 \times 4$ or $8 \times 8$ block is saved prior to partitioning the image to create the watermark. Fig. 5 depicts how $X_{w1}$ and $VMap_1$ are generated. Note that "Scale-up by 8" and "Scale-up by 4" indicate expansion of matrix elements into $8 \times 8$ blocks and $4 \times 4$ blocks, respectively.

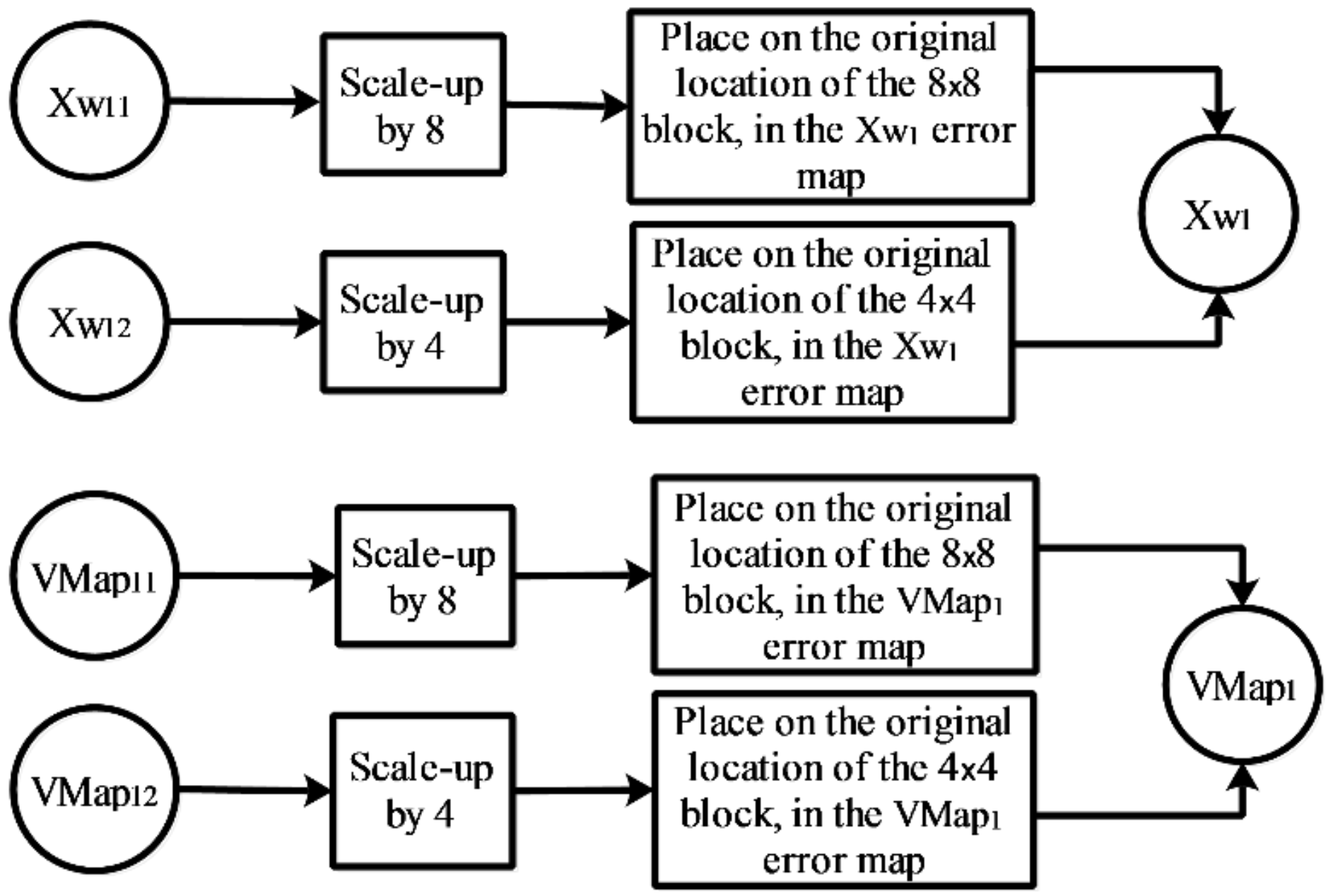

**Fig. 5** Constructing $X_{w1}$ and $VMap_1$

14- As explained, the elements in $X_{w21}$ and $VMap_{21}$ each represent an $8 \times 8$ block. Therefore, each element is expanded to an $8 \times 8$ block. While generating and extracting the second part of the watermark, $8 \times 8$ blocks are obtained via joining $4 \times 4$ blocks; in this step, a reverse operation occurs by breaking them (the expanded elements in $X_{w21}$ and $VMap_{21}$) down into $4 \times 4$ blocks. Next, given the fact that the elements in $X_{w22}$ and $VMap_{22}$ represent $4 \times 4$ blocks, each element is expanded to a $4 \times 4$ block. Considering the second stage of generating and extracting the watermark, each $4 \times 4$ block is part of a larger $8 \times 8$ block. Therefore, in this stage, four smaller blocks (that belong to an $8 \times 8$ block) are joined to represent an original $8 \times 8$ block. Once the final form of the $X_{w21}$, $X_{w22}$, $VMap_{21}$ and $VMap_{22}$ matrices are obtained, two $512 \times 512$ matrices namely $X_{w2}$ and $VMap_2$ are defined and initialized to zero. Then, in $X_{w2}$, the expanded elements from $X_{w21}$ and $X_{w22}$ are assigned to the locations of their respective representative blocks in the original image. Thus, we obtain the $X_{w2}$ error map. Similarly, $VMap_2$ is created using $VMap_{21}$ and $VMap_{22}$. The function of these matrices is described in a later section. The overall algorithm of creating $X_{w2}$ and $VMap_2$ can be seen in Fig. 6.

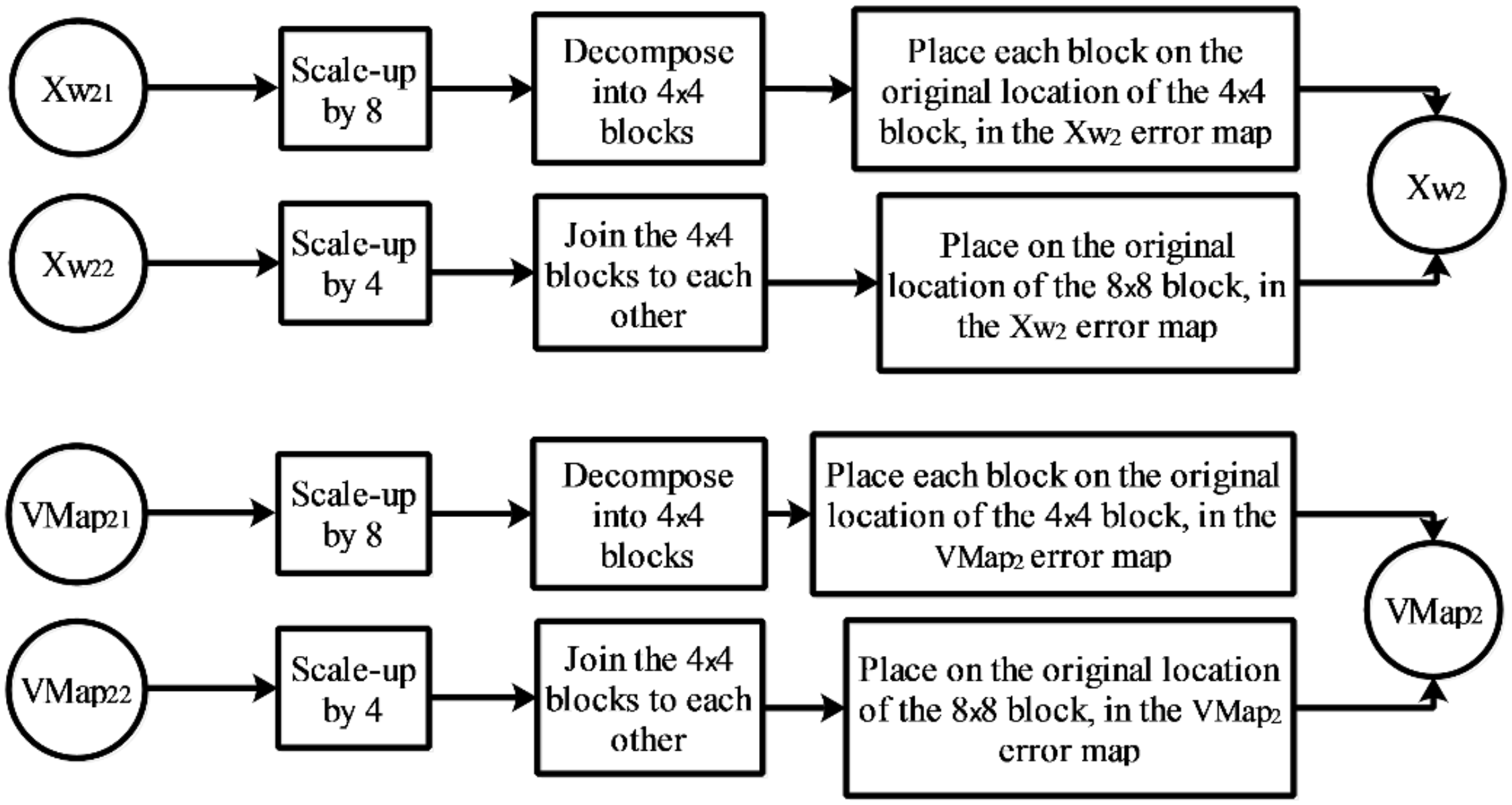

**Fig. 6** Constructing $X_{w2}$ and $VMap_2$

Thus, after these steps, $X_{w1}$, $X_{w2}$, $VMap_1$, and $VMap_2$ are obtained. The matrices may be regarded as four grayscale images. By displaying the $X_{w1}$ and $X_{w2}$ matrices as images and more precisely, by displaying the combination of them (the $X_{w1}$ and $X_{w2}$ matrices), the authenticity of an image can be ascertained. In other words, using the images or a combination thereof (obtained via Eq. 11), it is possible to determine whether the image has undergone tampering; the $X_{w1}$ and $X_{w2}$ matrices (images) identify the tampered region.

Moreover, in order to define a complementary numerical measure of authenticity, using the erosion and dilation operators, several features are computed and extracted from the error maps (previously described matrices) to classify the images. In fact:

15- Using a square window with $w = 5$, $X_{w1}$, $X_{w2}$, $VMap_1$, and $VMap_2$ undergo erosion, dilation, dilation, and erosion operations consecutively. Thereafter, Eq. 9 is used to calculate average pixel energy in the error map.

$$E(Error\ Map) = \frac{\sum_{i=1}^{N_R \times N_C} p_i^2}{N_R \times N_C}. \tag{9}$$

where $p_i$ is the $i$th pixel in the image; and $N_R \times N_C$ denotes image dimensions. Using Eq. 9, a total of eight values are obtained for the matrices; the values can be used as features to classify images into four classes which are as follows: (1) clean (watermarked) images, (2) images modified by image processing operations, (3) intentionally tampered images, and (4) images with both types of modification. Eq. 10, yields the eight aforementioned features. Note that $f_i$ , $i = \{1,2, \ldots ,8\}$ represents the feature computed from an error map while EDDE5 stands for erosion, dilation, dilation, and erosion using a window with $w = 5$.

$$\begin{aligned}
f_1 &= E(X_{w1}) \\
f_2 &= E(X_{w2}) \\
f_3 &= E(EDDE5(X_{w1})) \\
f_4 &= E(EDDE5(X_{w2})) \\
f_5 &= E(VMap_2) \\
f_6 &= E(EDDE5(VMap_2)) \\
f_7 &= E(VMap_1) \\
f_8 &= E(EDDE5(VMap_1))
\end{aligned} \tag{10}$$

Also, by combining $X_{w1}$ and $X_{w2}$ using Eq. 11, a new matrix $X_{w_comb}$ is obtained. By applying EDDE5 on this matrix, the energy of it can be determined and used as the ninth feature for classifying the images.

$$X_{w_comb} = \sqrt{X_{w1}^2 + X_{w2}^2} \tag{11}$$

where $X_{w1}^2 + X_{w2}^2$ is the sum of squared corresponding pixels in $X_{w1}$ and $X_{w2}$. Eq. 12 demonstrates how the ninth feature is obtained using the $X_{w_comb}$ error map (the map shown for each image in the Results section).

$$f_9 = E(EDDE5(X_{w_comb})) \tag{12}$$

Our experiments indicate that features $f_1$ through $f_6$ and also the feature $f_9$ are adequate for separating the images into different classes.

16- Finally, in order to classify the images into four different mentioned classes, a sufficient number of images are required so that the identified features can be calculated. Then a multi-class classifier is trained with the data so it can be used in the testing stage.

The next section discusses our results and compares them with those of the original scheme with respect to vulnerability.

## 3 Results and Comparison

In this section, the results obtained from several experiments are presented. The experiments are conducted using $512 \times 512$ grayscale images with $q = 8$. The following subsections present the results obtained from error maps and classification, respectively. It is noted that we compare our method with the method proposed in [17].

### *3.1 The Results Obtained Based on the Error Maps*

The error maps in this subsection are obtained from combining $X_{w1}$ and $X_{w2}$ according to Eq. 11. Furthermore, to enhance error map quality, the erosion and dilation operations ($w = 5$) are applied to $X_{w_comb}$.

In the first experiment (Fig. 7) neither intentional nor unintentional (image processing) attacks were present. As clear in the error maps, the error bits exhibit no distinct pattern. The small error present in the maps is caused by the fact that the watermark is generated using the image itself. Moreover, given the color of bits where errors occur, the errors are not significant.

It is noteworthy that we evaluated the performance of the proposed method and the method proposed in [17] under the PSNR value; It seems that the quality of the image, watermarked by using the proposed method, is better than the quality of the image, watermarked by using the method in [17]. Fig. 8 shows the comparison of PSNR values for fifty images, acquired for both proposed method and the method in [17].

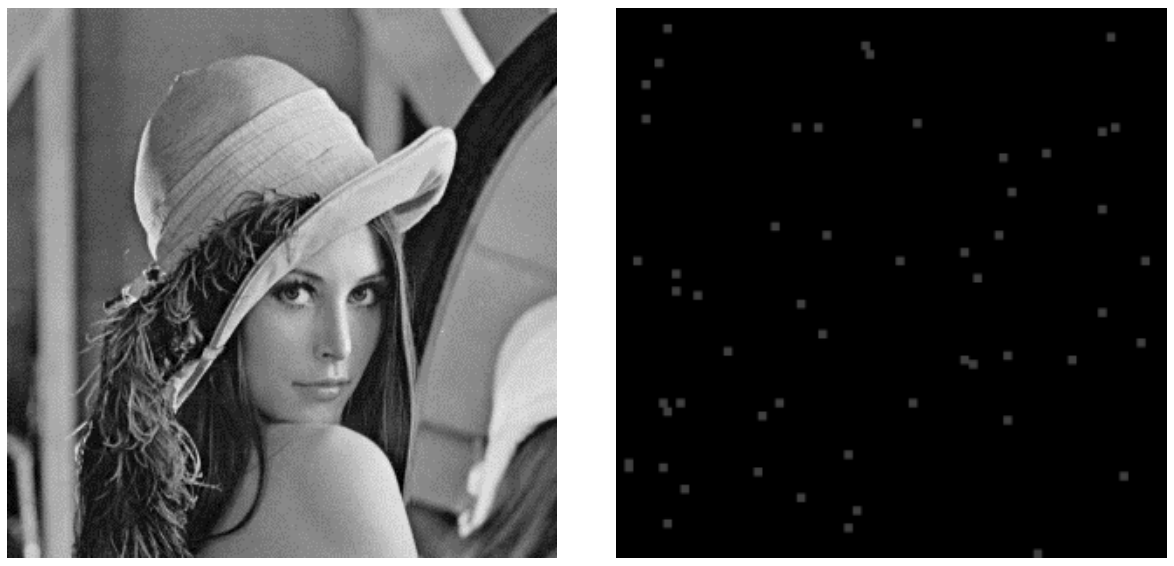

***(a)*** The watermarked image “Lena” with PSNR= 41.8912 dB and the error map resulting from authentication

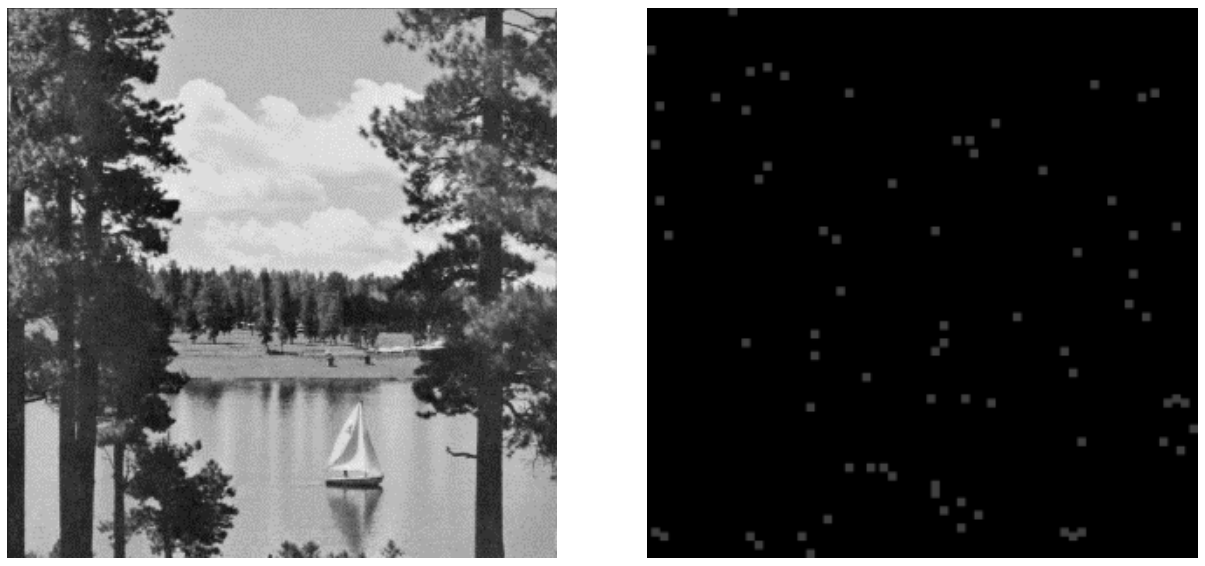

***(b)*** The watermarked image “Lake” with PSNR= 41.8275 dB and the error map resulting from authentication

**Fig. 7** Clean watermarked images (using the proposed scheme)

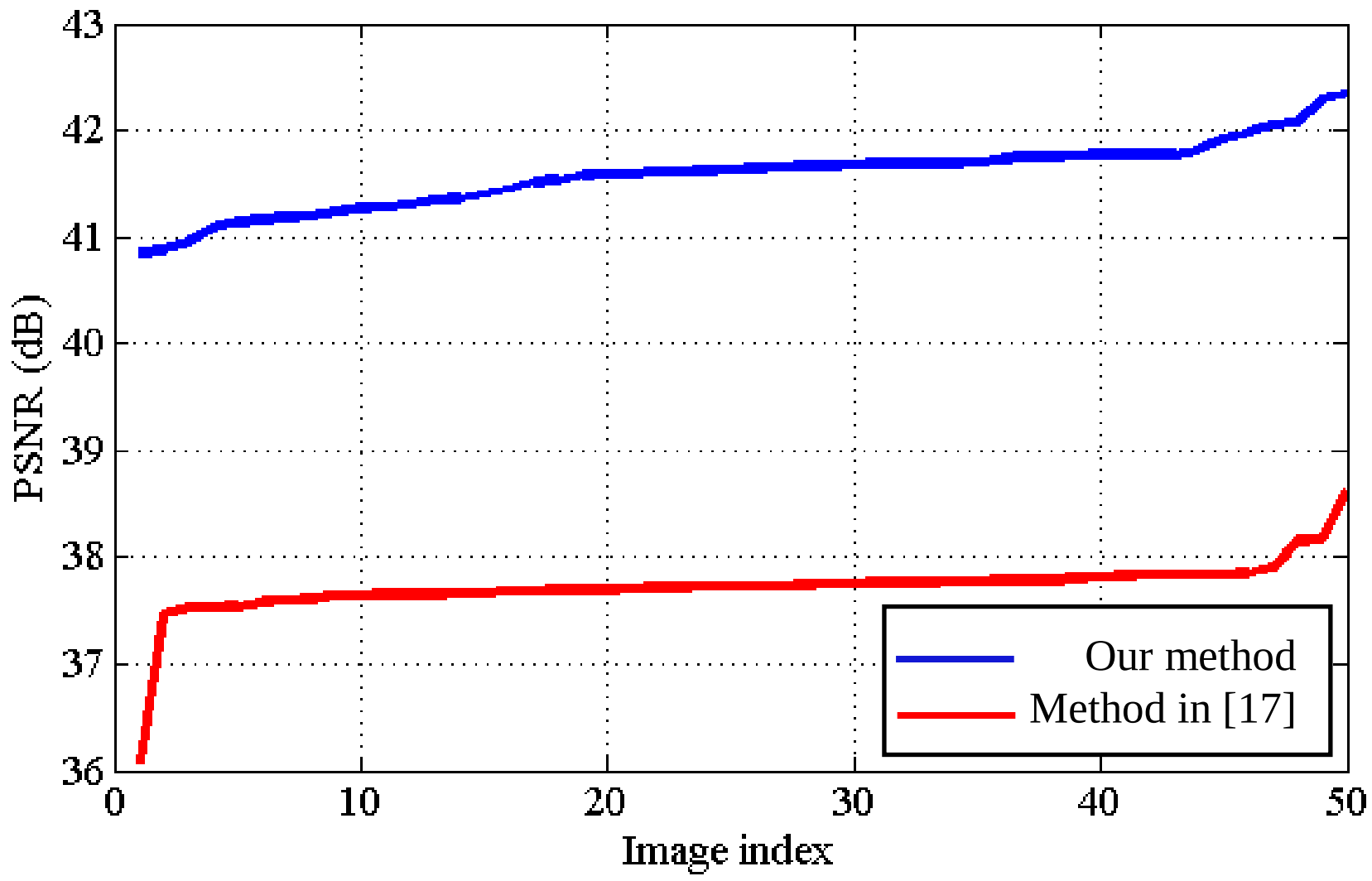

**Fig. 8** comparison of PSNR values for fifty images, acquired for both proposed method and the method in [17]

The purpose of the second experiment is to assess how robust the proposed method is against JPEG compression. Fig. 9 shows the results of applying JPEG compression on the image “Lena”, with QF= 75 to QF= 90 for both proposed method and the method proposed in [17]. As clear from the error maps (resulting from our authentication method), despite the increase in the extent of errors, a discernible pattern is not observed in the error maps; this is a sign of unintentional attacks. The authentication method proposed in [17] cannot distinguish the severity of distortion. In other words, in contrary to our proposed method, the authentication method in [17] does not make any difference between a region with high density tampered pixels and a region that has less tampered pixels. So it shows all the error regions with the pixels of white color in an error map. It is noted that the PSNR value obtained for the method in [17] is 37.7240 dB which represents less visual quality than our method (in terms of PSNR value).

The third experiment concerns fragility against intentional attacks. Here we consider an object insertion attack that inserts a piece of an unauthentic image into the watermarked image. The results of the attack on both proposed scheme and the scheme proposed in [17] are shown in Fig. 10. Both methods can detect the tampered regions. The error maps yielding from our proposed method consist of more noise pixels. But as their color shows, these errors are not so important and they are scattered over the error map. On the other hand, as mentioned earlier, by using the method proposed in [17], the watermarked image will have a significant decrease in visual quality.

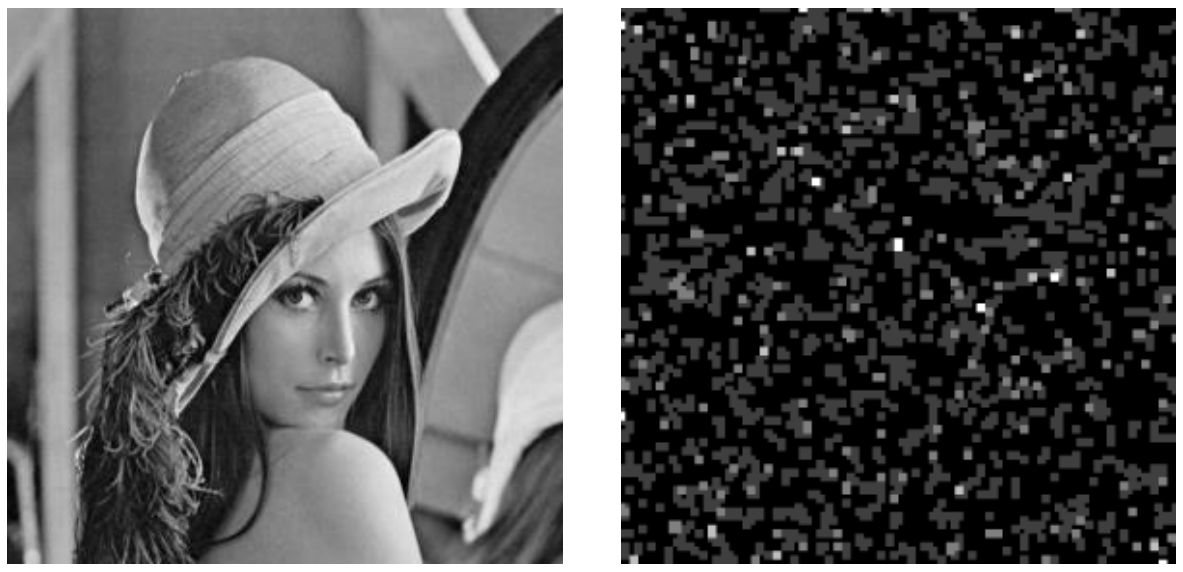

***(a)*** JPEG compression with QF =75 and the error map resulting from our proposed authentication method

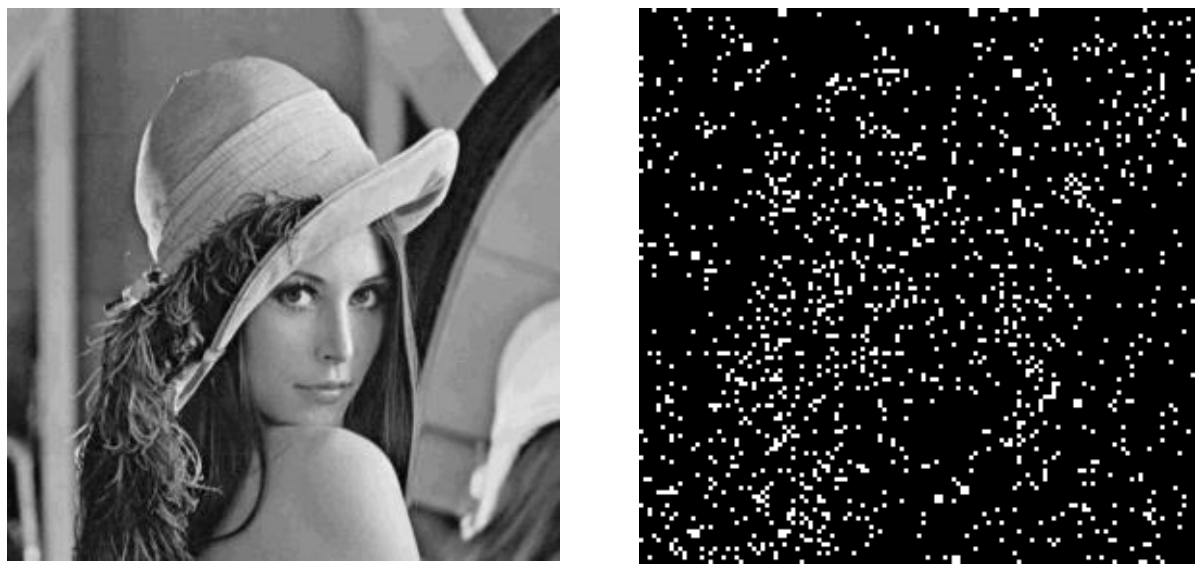

***(b)*** JPEG compression with QF = 75 and the error map resulting from the authentication method proposed in [17]

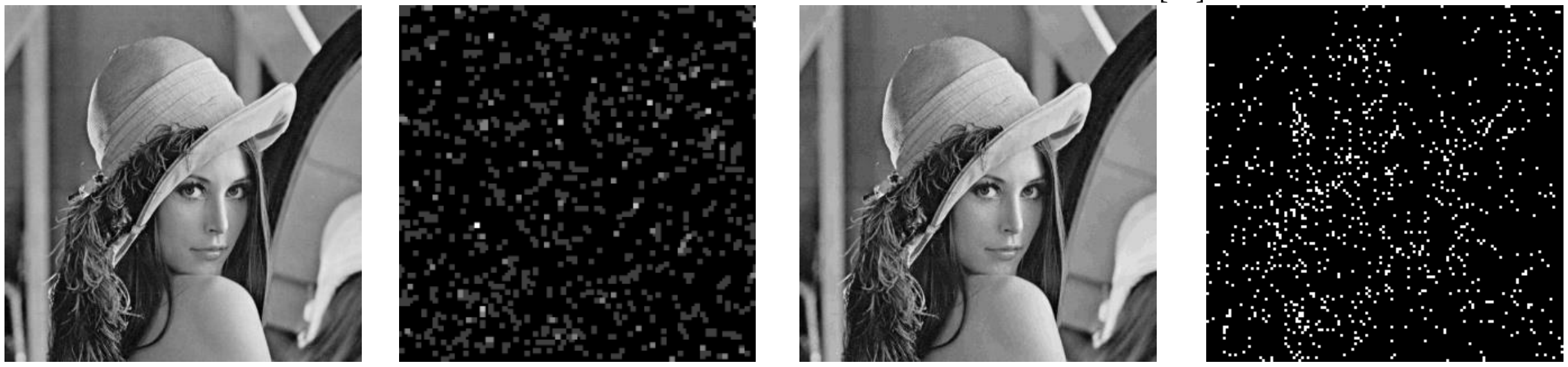

***(c)*** JPEG compression with QF = 80 and the error map resulting from our proposed authentication method

***(d)*** JPEG compression with QF = 80 and the error map resulting from the authentication method proposed in [17]

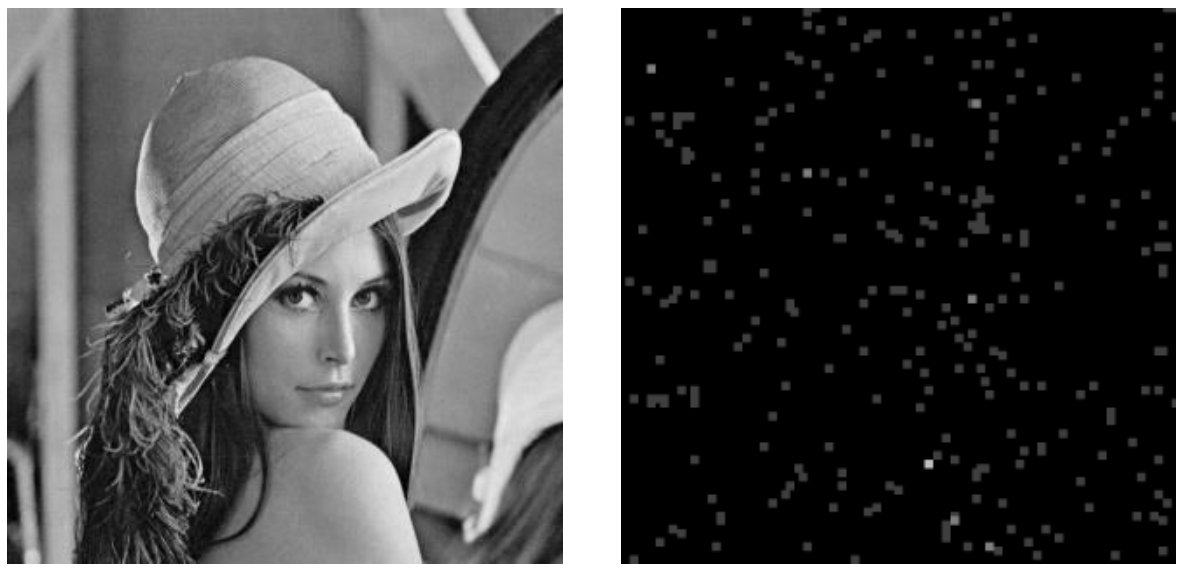

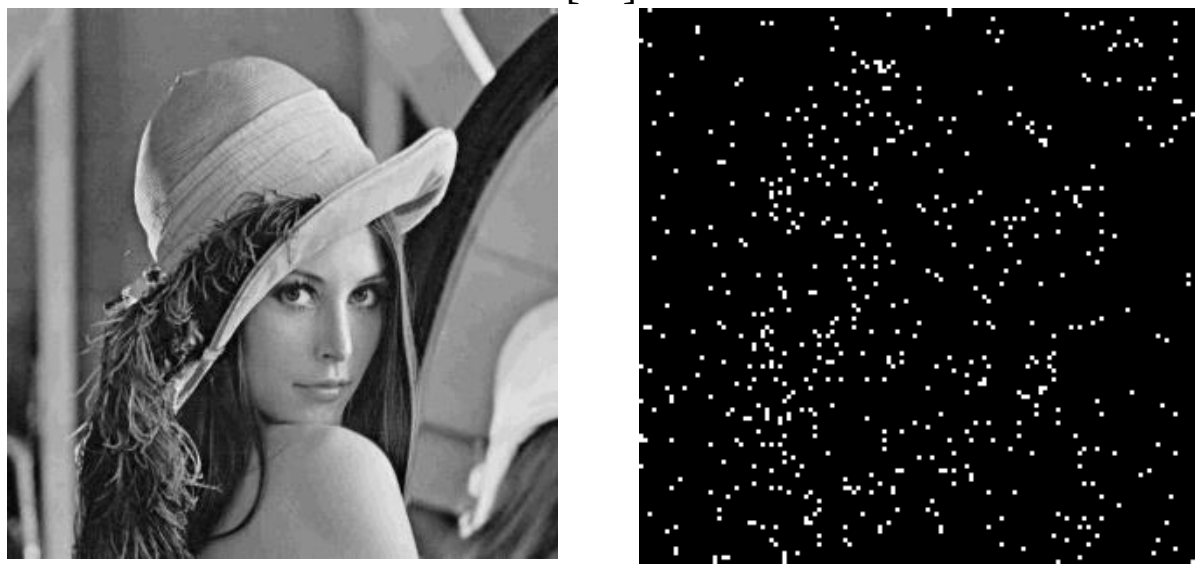

***(e)*** JPEG compression with QF = 85 and the error map resulting from our proposed authentication method

***(f)*** JPEG compression with QF = 85 and the error map resulting from the authentication method proposed in [17]

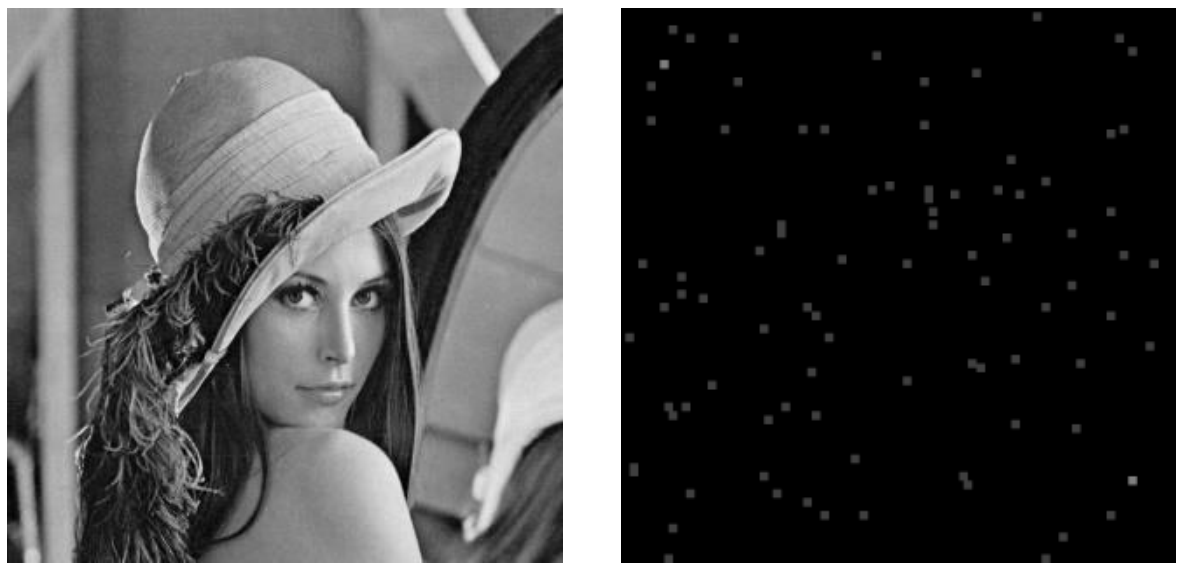

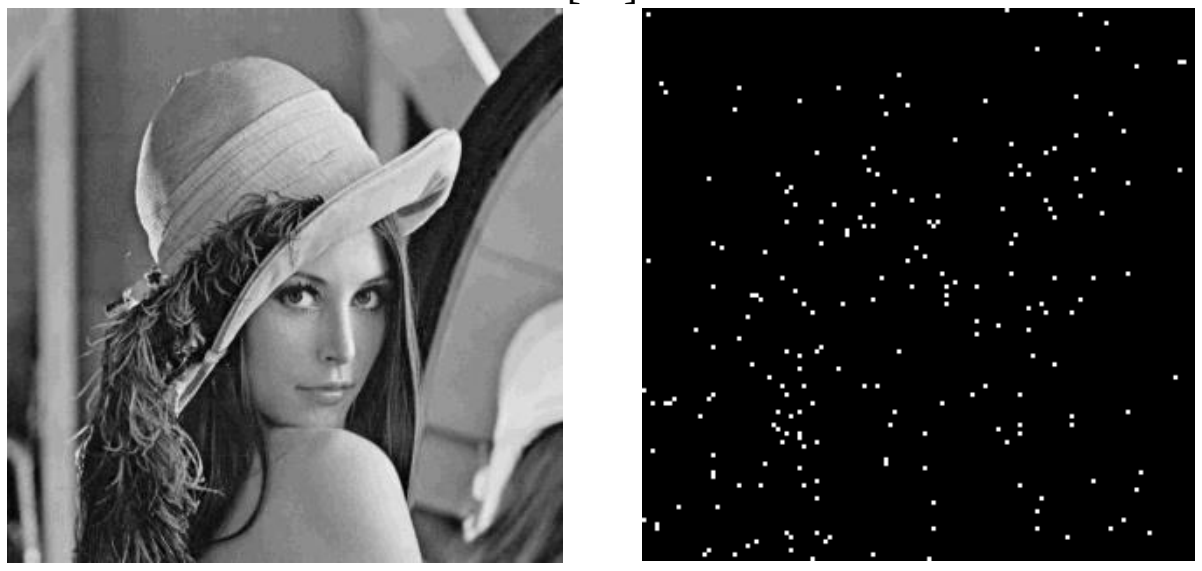

***(g)*** JPEG compression with QF =90 and the error map resulting from our proposed authentication method

***(h)*** JPEG compression with QF =90 and the error map resulting from the authentication method proposed in [17]

**Fig. 9** "Lena" watermarked image subsequent to JPEG compression

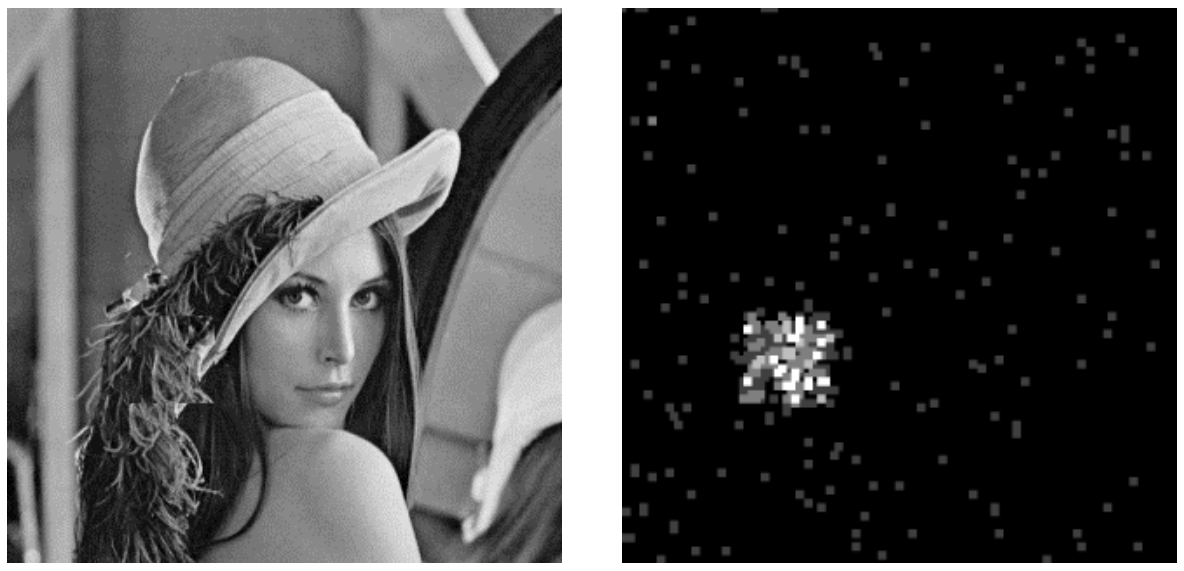

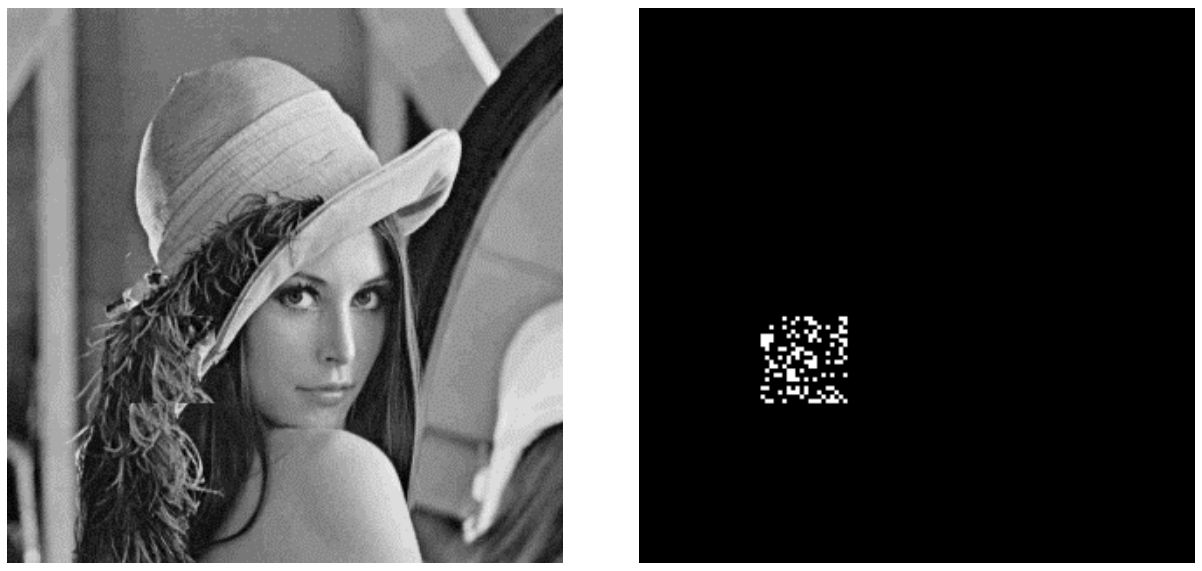

***(a)*** Tampered "Lena" image, watermarked using the proposed method and the error map resulting from the proposed authentication strategy

***(b)*** Tampered "Lena" image, watermarked using the method proposed in [17] and the error map resulting from it's authentication strategy

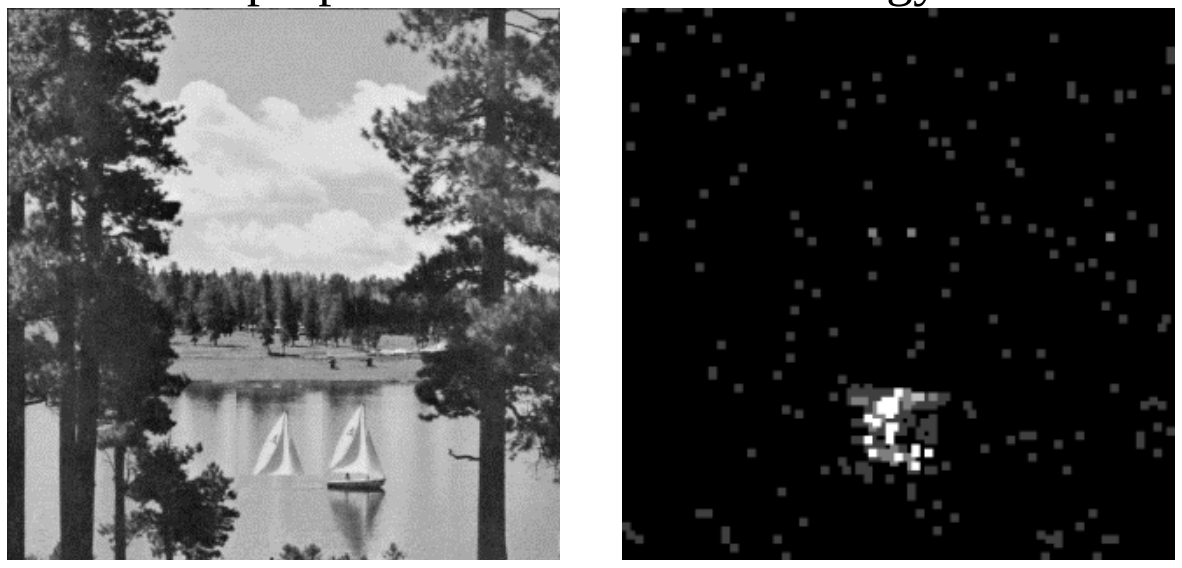

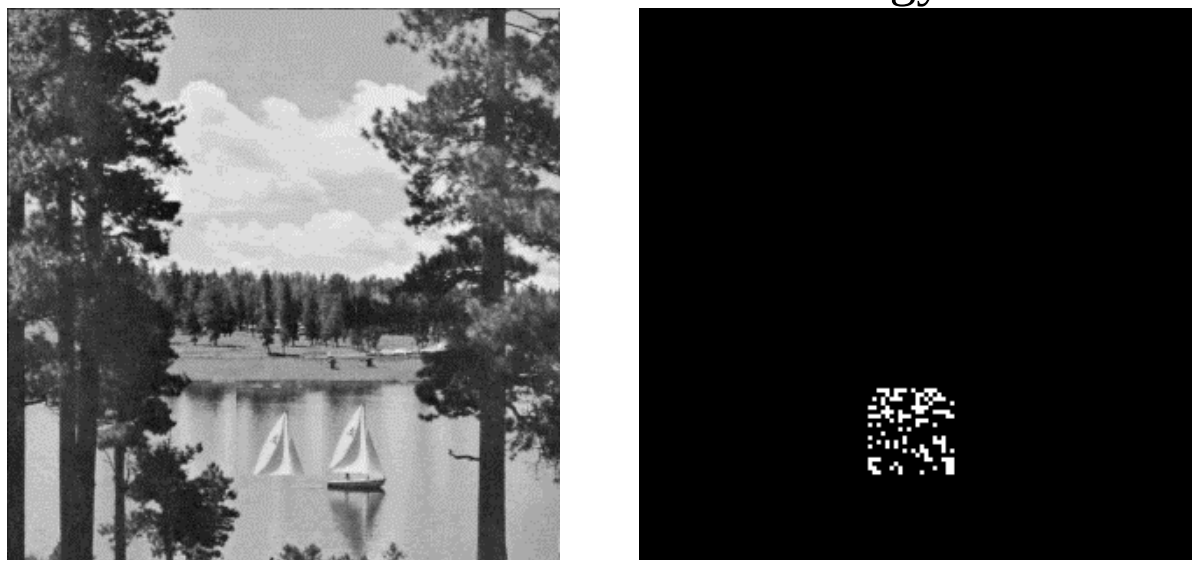

***(c)*** Tampered "Lake" image, watermarked using the proposed method and the error map resulting from the proposed authentication strategy

***(d)*** Tampered "Lake" image, watermarked using the method proposed in [17] and the error map resulting from it's authentication strategy

**Fig. 10** Results of applying the object insertion attack on both schemes

Finally, the last experiment examined the watermarked image (watermarked using the proposed method), with both intentional and unintentional attacks. Indeed, an intentional attack was followed by JPEG compression. The results indicate that the proposed method is robust against JPEG compression while being sensitive to intentional attacks. With regards to the error maps, the error bits are scattered in the former case (JPEG compression) as opposed to the concentrated pattern observed in the latter case (intentional attacks). In Fig. 11, the proposed method and the method proposed in [17] are subjected to the object insertion attack together with JPEG compression where QF equals 75 to 90. As it is clear, by decreasing the JPEG quality factor, both methods show more noise pixels in the error maps. But the noise pixels resulted from our proposed authentication method have colors which are close to black. In the intentionally tampered regions, the error pixels in our error maps have colors that are close to white. Fig. 11(b) shows that the method proposed in [17] cannot distinguish the intentionally tampered regions from the scattered noise pixels which are resulted from JPEG compression.

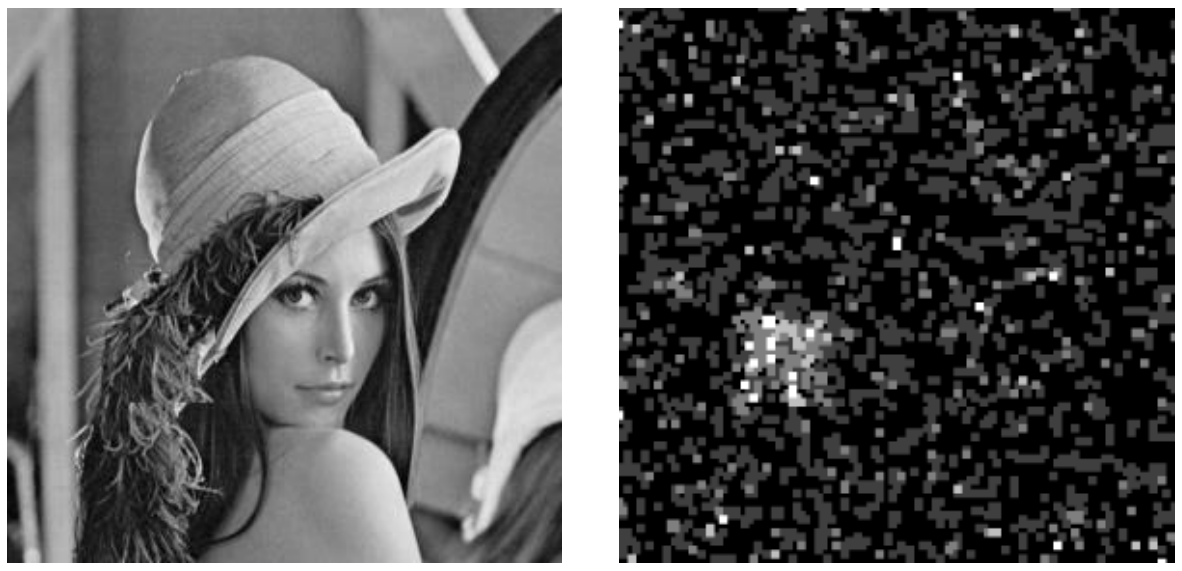

***(a)*** the watermarked "Lena" image subsequent to the object insertion attack and JPEG compression with QF = 75, and the error map resulting from proposed image authentication

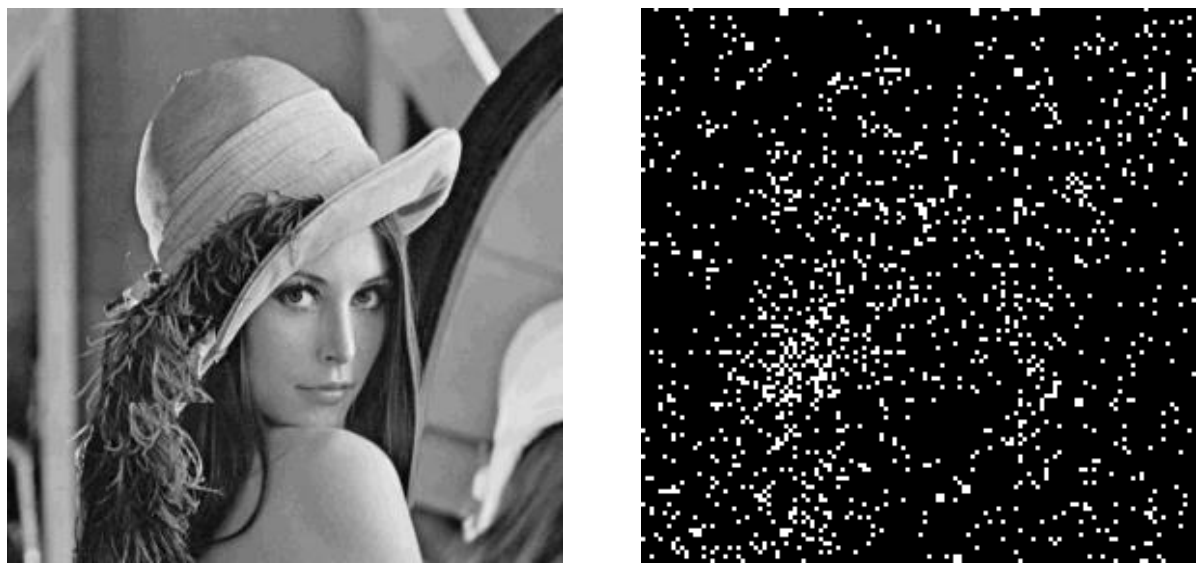

***(b)*** the watermarked "Lena" image subsequent to the object insertion attack and JPEG compression with QF = 75, and the error map resulting from image authentication proposed in [17]

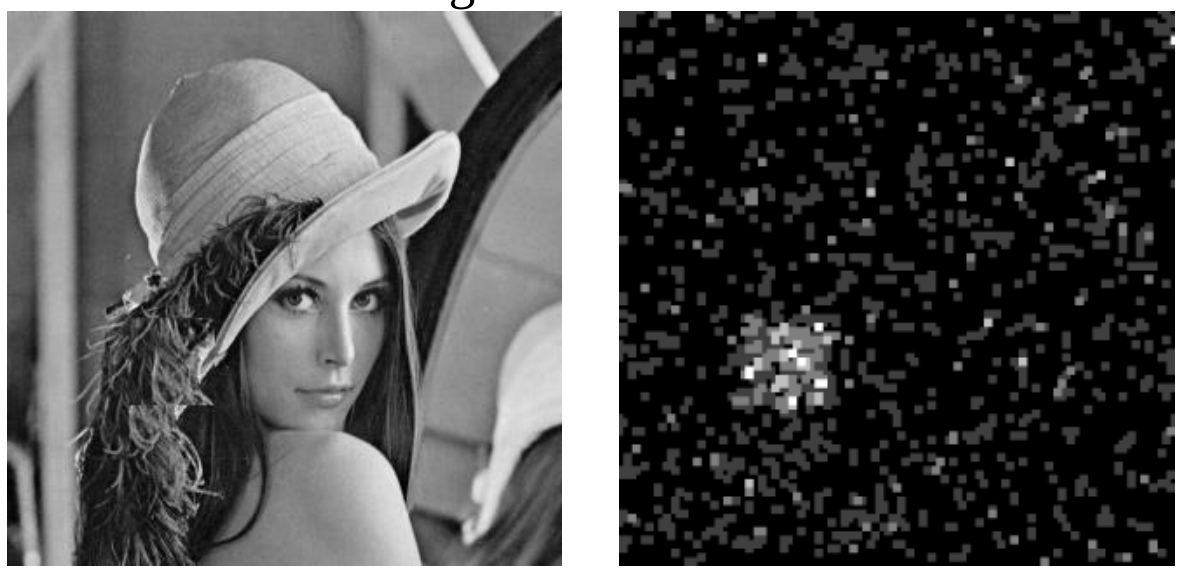

***(c)*** the watermarked "Lena" image subsequent to the object insertion attack and JPEG compression with

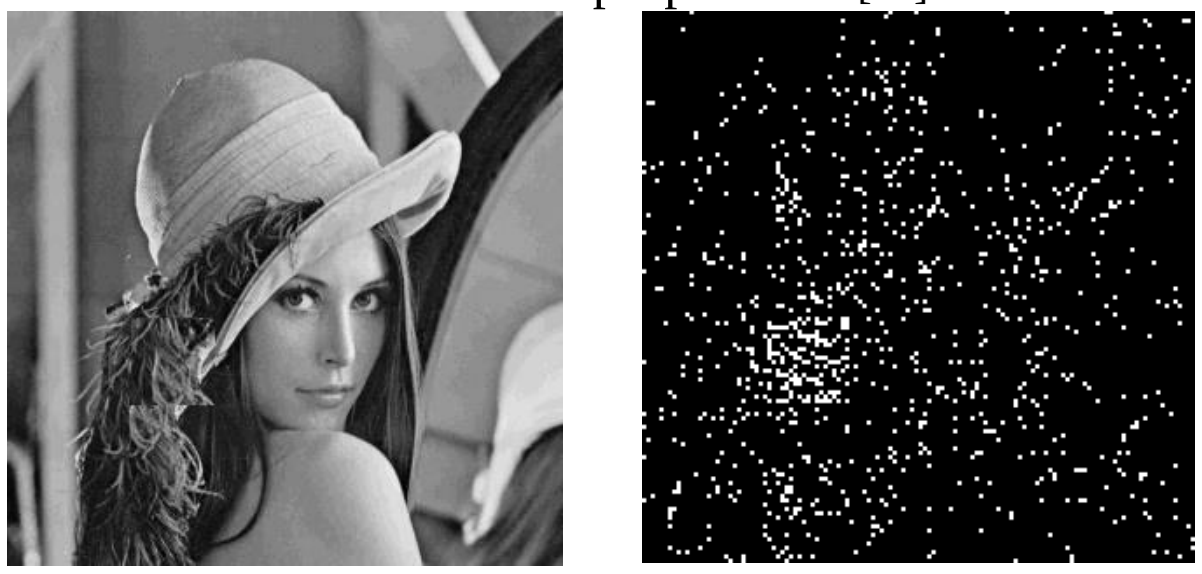

***(d)*** the watermarked "Lena" image subsequent to the object insertion attack and JPEG compression with QF

QF = 80, and the error map resulting from proposed image authentication

= 80, and the error map resulting from image authentication proposed in [17]

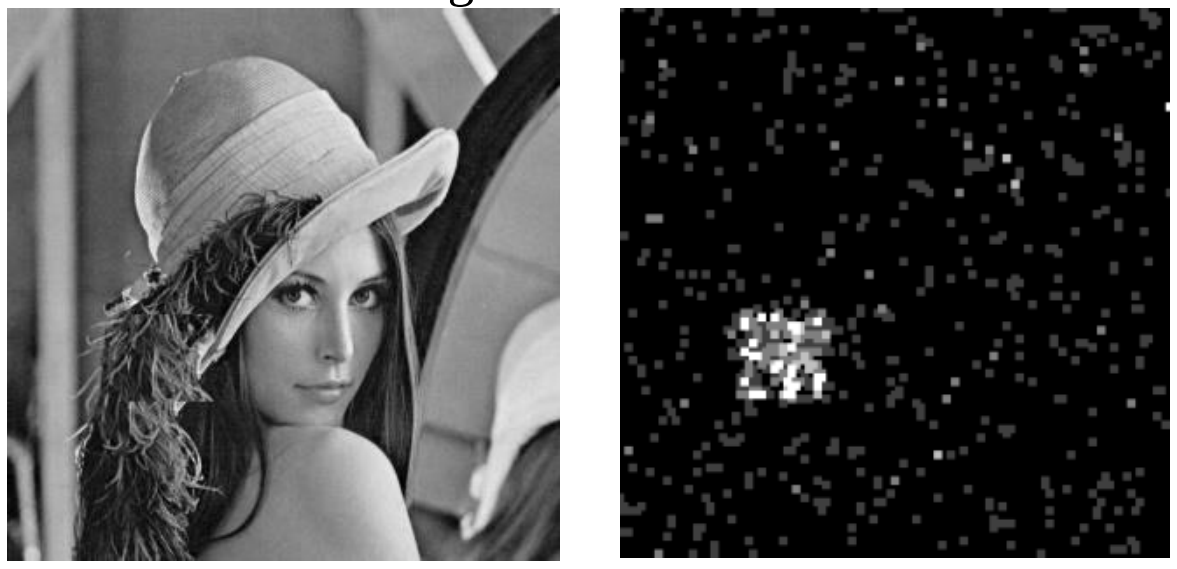

***(e)*** the watermarked “Lena” image subsequent to the object insertion attack and JPEG compression with QF = 85, and the error map resulting from proposed image authentication

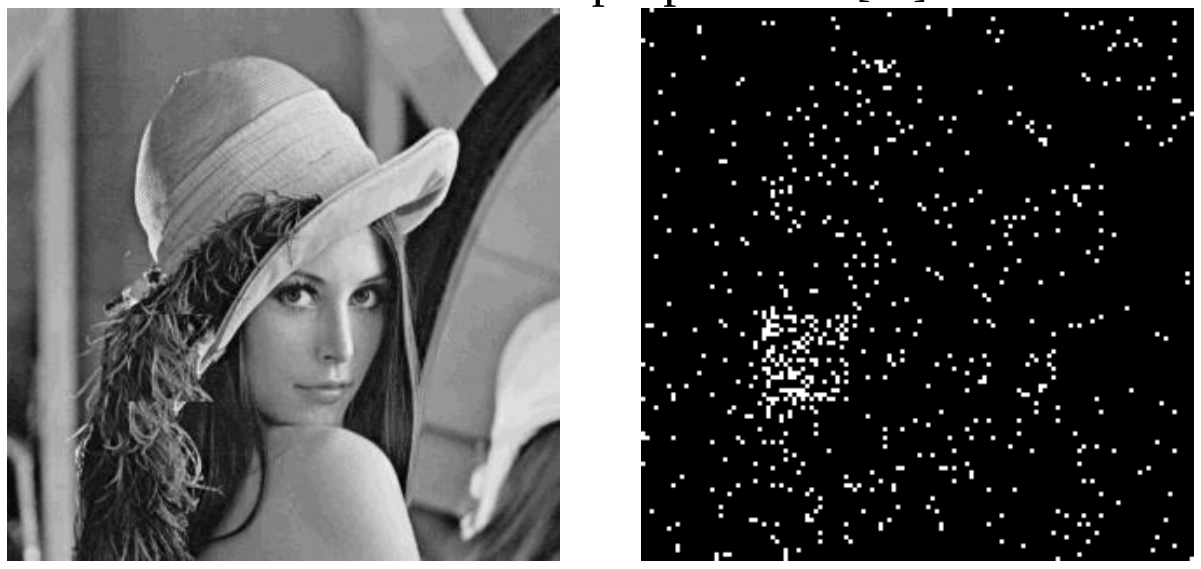

***(f)*** the watermarked “Lena” image subsequent to the object insertion attack and JPEG compression with QF = 85, and the error map resulting from image authentication proposed in [17]

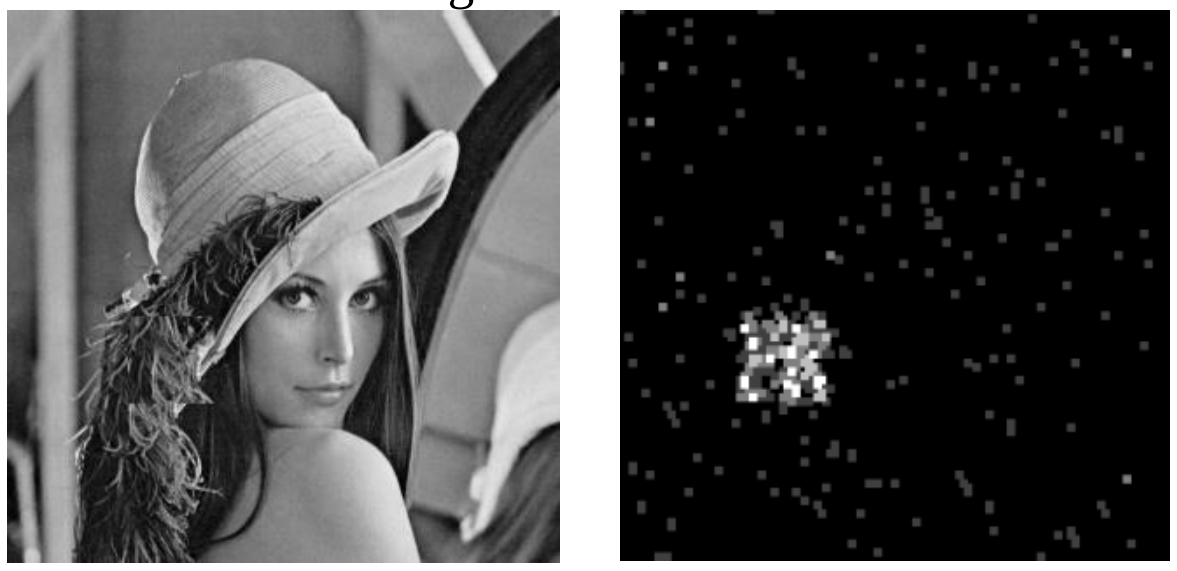

***(g)*** the watermarked “Lena” image subsequent to the object insertion attack and JPEG compression with QF = 90, and the error map resulting from proposed image authentication

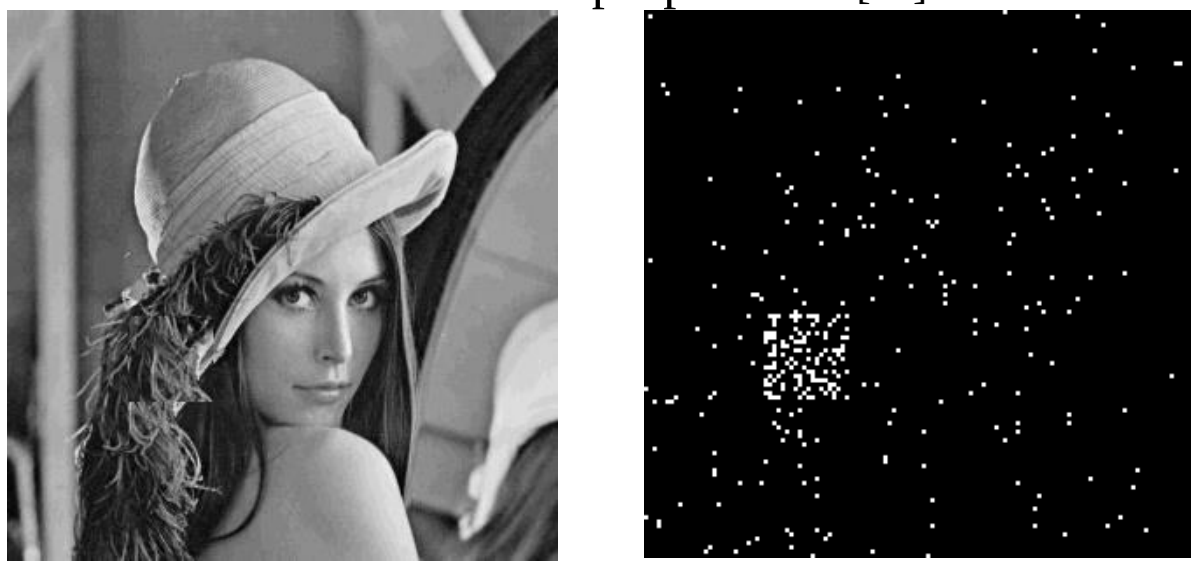

***(h)*** the watermarked “Lena” image subsequent to the object insertion attack and JPEG compression with QF = 90, and the error map resulting from image authentication proposed in [17]

**Fig. 11** Results of applying an intentional attack together with an unintentional attack

As the method in [17], the time complexity of the proposed method is linear with the size of image. Table 4 shows the average time of watermark embedding and watermark extraction for a 512× 512 grayscale image, for both proposed scheme and the method proposed in [17]. As it is clear, the proposed method can embed and extract the watermark in a less noticeable time than the method in [17].

**Table 4** Comparing the average time of embedding

| Watermarking method | Average embedding time for an image (in seconds) | Average extraction time for an image (in seconds) |
|---|---|---|
| Proposed method | 82.4923 | 41.5057 |
| [17] | 155.9872 | 61.8467 |

Now, the images must be classified using features $f_1$ through $f_6$ and also the feature $f_9$ to verify the accuracy of tamper detection and to ensure the correctness of classification.

### *3.2 The Results of Classification*

For classification purposes, a total of four categories were considered with 600, 1800, 1200, and 3600 grayscale images of $512 \times 512$ pixels and the features were calculated. It should be noted that the images were selected from the SUN dataset [26]. The results were subsequently used to train and test a classifier. The categories are as follows:

1- Clean watermarked images with neither intentional nor unintentional attacks. Also, JPEG compressed images with QF = 100 are assigned to this category given their slight difference with clean images (600 images).

2- Watermarked images having only JPEG compression (1800 images) where QF = 75 to 95 (no intentional attacks are present).

3- Watermarked images with an intentional attack which may have undergone JPEG compression with QF = 100 (1200 images).

4- Watermarked images with both an intentional attack and JPEG compression where QF = 75 to 95 (3600 images).

To classify the images, the features obtained from the instances (images) along with the label for each instance are given as input to a multi-class SVM (with RBF kernel type) in RapidMiner 5.2. Thus, data normalization, training and test processes are performed in a standard manner to yield overall classification accuracy as well as the accuracy for individual categories. The validation process in the training and test phases uses 1 against all classification with 15 subsets for the cross validation.

Table 5 presents the classification results using the aforementioned seven features. In addition to overall accuracy, the values at the bottom of each column represent detection accuracy for a specific category. More specifically, $true\ x$ is the percentage of correctly classified images while $pred.x$ is the percentage of images with the label $x$. Each column shows the number of correctly (darker cell) and incorrectly classified images for the respective category.

**Table 5** Classification results using the seven features

| Accuracy: 97.97% +/- 0.69% | | | | | |
|---|---|---|---|---|---|
| | True Clean or Jp100 (true 1) | True Jp75 or Jp80 or Jp85 or Jp 90 or Jp95 (true 2) | True Attack or Attack+Jp100 (true 3) | True Attack+Jp75 or Jp80 or Jp85 or Jp90 or Jp95 (true 4) | Class precision |
| Pred.Clean or Jp100 (Pred.1) | 594 | 0 | 0 | 0 | 100.00% |
| Pred. Jp75 or Jp80 or Jp85 or Jp90 or Jp95 (Pred.2) | 0 | 1710 | 0 | 10 | 99.42% |
| Pred. Attack or Attack+Jp100 (Pred.3) | 6 | 0 | 1180 | 20 | 97.84% |
| Pred.Attack+ Jp75 or Jp80 or Jp85 or Jp90 or Jp95 (Pred.4) | 0 | 90 | 20 | 3570 | 97.01% |
| class recall | 99.00% | 95.00% | 98.33% | 99.17% | |

In this section, authentication and classification results were presented for images watermarked using the proposed approach. In addition to the proposed error maps that are able to correctly distinguish between the types of attacks, high accuracy (97.97 percent) is achieved for classification which indicates the usability of the proposed features. It is noted that the method proposed in [17] does not have any classification mechanism to classify the types of attacks.

## 4 Conclusion

Semi-fragile watermarking methods are comparatively more efficient than fragile and robust methods. Therefore, semi-fragile methods can be trusted to replace paper-based mechanisms in authenticating critical documents if the aforementioned objective is achieved in practice. Thus, this paper is a try to achieve this goal by proposing a semi-fragile watermarking method.

In this paper, a watermarking method was presented that is mainly characterized by the followings: (1) the watermark is generated using image content i.e. $8 \times 8$ and $4 \times 4$ random blocks; (2) the watermarked blocks are dependent to each other; (3) tampered regions are identified visually using an error map; and (4) authentication is achieved based on the features, calculated using the error maps. More precisely, in authentication stage, in addition to displaying the error maps, average pixel energy of the images (error maps) is suitable in classifying the images into four categories, including: (1) clean watermarked images, (2) unintentionally modified images, (3) intentionally tampered images, and (4) images with both intentional and unintentional modifications. Classification was performed using SVM in RapidMiner and achieved an accuracy of 97.97 percent.

As clear in the error maps and classification results, the proposed method is able to correctly detect intentional, unintentional and the combination of both attacks. We also showed that our method outperforms a state-of-the-art semi-fragile watermarking method. It should be noted that, by making the watermark dependent on image content, the method becomes slightly less robust and more sensitive to unintentional attacks. We showed that

**Funding:** This research did not receive any specific grant from funding agencies in the public, commercial, or not-for-profit sectors.

**Samira Hosseini** received the B.S. in IT Engineering from University of Isfahan, Isfahan, Iran, in 2014 and received the M.S. in Information Security at University of Isfahan, in 2017. Her current research interests include Watermarking and Steganalysis.

**Mojtaba Mahdavi** received the B.S. in Computer hardware Engineering From Isfahan University of technology, Iran in 1999. He received the M.S. in Computer architecture from the Isfahan University of Technology in 2002 and his Ph.D. in Electrical Engineering from Isfahan University of Technology, in 2011. He is now an Assistant Professor in the Department of Information Technology, University of Isfahan. His current research interests include Steganography, Steganalysis, Watermarking and Network Covert Channels.